\documentclass{article}
\usepackage{iclr2027_conference,times}

\usepackage{amsmath,amsfonts,bm}

\def\eqref#1{equation~\ref{#1}}
\def\1{\bm{1}}

\DeclareMathAlphabet{\mathsfit}{\encodingdefault}{\sfdefault}{m}{sl}
\SetMathAlphabet{\mathsfit}{bold}{\encodingdefault}{\sfdefault}{bx}{n}

\usepackage{hyperref}
\hypersetup{hidelinks,pdftitle={QRAE: Query-Relevant Attention Entropy for Context Compression in DeepSearch Agents},pdfauthor={Yingjie Tan, Haiquan Qiu, Quanming Yao, Manyi Zhong, Xianzhi Yu, Yaqing Wang}}
\usepackage{url}
\usepackage{graphicx}
\usepackage{subcaption}
\usepackage{multirow}
\usepackage{booktabs}
\usepackage{algorithm}
\usepackage{algpseudocode}
\usepackage{xcolor}
\usepackage{amsmath}
\usepackage{amssymb}
\usepackage{xspace}
\usepackage{enumitem}
\usepackage{array}
\usepackage{amsthm}
\usepackage{float}
\usepackage{microtype}

\newcommand{\MethodName}{QRAE\xspace}
\newcommand{\SignalName}{\MethodName}
\newcommand{\EAH}{evidence-assessment head\xspace}
\newcommand{\EAHs}{evidence-assessment heads\xspace}
\newcommand{\Attention}{\operatorname{Attention}}
\newcommand{\Entropy}{\operatorname{Entropy}}
\newtheorem{proposition}{Proposition}[section]
\newtheorem{lemma}[proposition]{Lemma}

\title{QRAE: Query-Relevant Attention Entropy\\
for Context Compression in\\
DeepSearch Agents}

\iclrfinalcopy
\author{%
\textbf{Yingjie Tan$^{1}$\thanks{Core-author contact: Tan Yingjie, \href{mailto:tanyj23@mails.tsinghua.edu.cn}{\texttt{tanyj23@mails.tsinghua.edu.cn}}.} \quad Haiquan Qiu$^{1}$ \quad Quanming Yao$^{1}$}\\
\textbf{Manyi Zhong$^{2}$ \quad Xianzhi Yu$^{2}$ \quad Yaqing Wang$^{1}$}\\[0.5em]
{\normalfont $^{1}$Tsinghua University \quad $^{2}$Huawei Technologies Ltd.}%
}

\begin{document}

\maketitle
% Remove the conference-publication header for this preprint.
\fancyhead{}
\renewcommand{\headrulewidth}{0pt}
% Balance title spacing after adding the author-contact footnote.
\vspace{-12pt}

\begin{abstract}
DeepSearch LLM agents repeatedly search and carry retrieved sources into
subsequent reasoning steps. This expanding context increases token
consumption and latency, although many results do not contribute to later
reasoning. We propose \textbf{Query-Relevant Attention Entropy (QRAE)}, an online
context compression method that uses sparse evidence-assessment heads to decide
which intact retrieved chunks should persist across reasoning steps.
It scores bounded candidate batches through paired prefills: a query-free control
suppresses shared attention responses, while normalized entropy measures the breadth of
the positive response within each chunk. High-scoring chunks enter a persistent
evidence whitelist and are included in the reasoning context. Uncertain chunks
remain outside the reasoning context but are retained in a candidate cache,
where they can be retrieved for later reevaluation. \MethodName requires neither
an additional generative compression pass nor modification of the attention operator.
% Macro-averages over the six model--benchmark settings in Tables 1 and 2:
% success-rate differences and within-setting relative latency/token reductions vs. DeepSearch.
% Token accounting follows the respective tables (input tokens for OpenPangu).
Across multiple DeepSearch benchmarks and reasoning models,
\MethodName improves task success by an average of 2.35 percentage points and
reduces end-to-end latency and reported token usage by an average of 30\% and
45\%, respectively. These benefits generalize to a model with nearly 100B parameters.
\end{abstract}

\section{Introduction}
\label{sec:introduction}

DeepSearch agents answer information-intensive questions through a cycle of
thought and action, as exemplified by Tongyi DeepResearch and
OpenResearcher~\citep{tongyi2025deepresearch,li2026openresearcher}. At each step,
the agent generates reasoning and a tool action, such as searching, opening a
page, or finding information within it, then reads the result and continues
until it can synthesize an answer. Search returns text snippets, also known as
\emph{chunks}. Because each search returns multiple results, these snippets
make up most of the reasoning context and accumulate across steps, requiring
repeated processing in later model calls.

To reduce token cost and bound context length, many agent systems invoke an
LLM to summarize, reflect on, or refine the accumulated history. RE-TRAC turns
completed search trajectories into structured states for subsequent
exploration~\citep{zhu2026retrac}; ACON condenses observations and interaction
histories; and Context-Folding replaces completed subtrajectories with compact
summaries~\citep{kang2026acon,sun2025contextfolding}. Although effective, these
approaches require an additional language-model compression pass, adding
compute and wall-clock latency. Retrieval compression can instead prune
passages before reasoning, while KV-cache compression reduces serving memory
and attention cost~\citep{sharifymoghaddam2026rerank,li2024snapkv,
zhang2023h2o}. These methods remain useful, but they operate on different
dimensions from deciding which redundant search results should persist across
reasoning steps. This motivates a mechanism-grounded compression method that
derives its selection signal from the reasoning model itself without an
additional LLM call.

We propose \textbf{Query-Relevant Attention Entropy (QRAE)}, an online context
compression method that scores search snippets before they enter the reasoning
context, using attention entropy from the reasoning model itself. New chunks
enter a candidate cache, and each
bounded candidate batch receives normalized-entropy scores from prefill-only
evaluation.
Chunks ranked highest by their running-mean scores enter a persistent evidence whitelist and
remain in the reasoning context across subsequent steps. Uncertain chunks stay
in a candidate cache outside the reasoning context, allowing the manager to
retrieve, reevaluate, and admit them when later evidence changes their utility.

We use the attention entropy of a small set of localized query-relevant heads
as a model-internal signal of each search snippet's relevance to the current
query. We draw inspiration from transformer-circuit
analyses~\citep{elhage2021mathematical,olsson2022context}: models contain circuits
that direct attention toward query-relevant input tokens before answer
generation, acting as an internal gating mechanism. We
therefore localize heads offline whose query-conditioned prefill attention
consistently responds to answer-bearing spans. We observe that relevant
chunks tend to elicit responses across multiple supporting tokens, whereas
incidental matches produce isolated peaks. We use normalized entropy to
quantify the breadth of the positive query-induced response, yielding the
\SignalName score used by the compression framework.

We evaluate \MethodName across multiple models and datasets. End-to-end agent
experiments show that \MethodName substantially reduces wall-clock time and
token consumption while modestly improving overall task success. The
\SignalName signal identifies relevant chunks containing supporting evidence.
Sparse query-relevant heads recur across models and architectures, including
hybrid architectures with linear-time components and sparse-attention models.
Their consistent responses across datasets and reasoning steps support a
model-level mechanism that generalizes beyond the localization data. Finally,
strengthening the signals of these heads substantially improves downstream
task performance.

Our contributions are:
\begin{itemize}[leftmargin=*,nosep]
    \item \textbf{Method contribution:} We propose a context compression method for DeepSearch agents,
    framing context management as decisions about which intact retrieved
    snippets to pass to the reasoning context. It combines model-internal
    scoring with recoverable deferral to manage context efficiently.
    \item \textbf{Mechanistic contribution:} Through mechanistic analysis, we identify a model-internal evidence
    signal carried by a family of model-level heads. These heads consistently
    respond to query-relevant input, and this mechanism generalizes across
    models of different sizes and architectures.
    \item \textbf{Experimental contribution:} Agent experiments show that \MethodName substantially reduces end-to-end
    time and token costs while modestly improving overall task success.
    Evaluation experiments show that the \SignalName signal effectively identifies
    query-relevant input content.
\end{itemize}

\section{QRAE Agent Framework}
\label{sec:framework}

\begin{figure}[t]
\centering
\includegraphics[width=\textwidth]{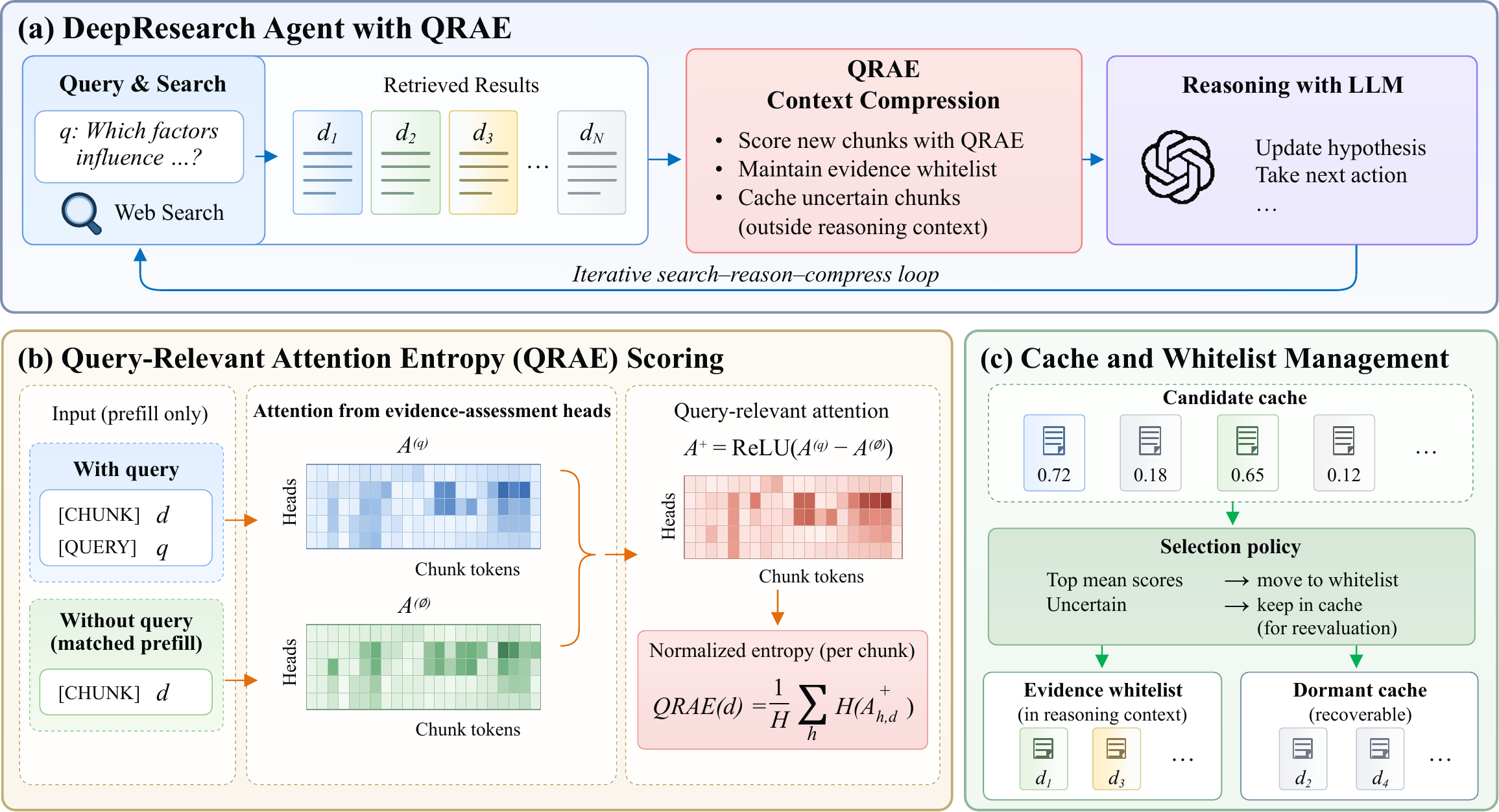}
\caption{Overview of \MethodName for context compression in DeepSearch
agents. (a) Retrieved chunks pass through \MethodName before the LLM updates its
hypothesis and takes the next action. (b) The \SignalName signal is derived
from the attention distributions of the model's query-relevant heads.
(c) Chunks with the highest running-mean scores in the candidate cache enter
the whitelist and the reasoning context; deferred chunks remain in the
candidate cache.}
\label{fig:qrae_context_flow_framework}
\end{figure}

This section describes how the \MethodName framework manages agent context.
It uses model-internal attention entropy scores to select retrieved chunks
for the reasoning context, reducing the length of context supplied to the
model and enabling efficient inference.
Figure~\ref{fig:qrae_context_flow_framework} summarizes how the \MethodName
framework interacts with the reasoning LLM. The construction of the
\SignalName signal is detailed in Section~\ref{sec:qrae-signal}.

\subsection{Problem Formulation}
\label{sec:problem-formulation}

At agent step $t$, the DeepSearch agent receives a reasoning context $P_t$ and
generates reasoning followed by a tool action. A search action returns snippets
$\mathcal{D}_t=\{d_{t,1},\ldots,d_{t,k}\}$, each of which is one \emph{chunk}.
These snippets form the
bulk of the reasoning context. Without compression, they
accumulate in the interaction history and are repeatedly processed by later
model calls, even when redundant or no longer useful. \MethodName compresses
this portion by scoring and selecting snippets before they enter the reasoning
context.

The reasoning context is the input supplied to the agent for a reasoning step.
The context manager maintains two disjoint states. The \emph{persistent
whitelist} $\mathcal{W}_t$ contains admitted chunks that are included in the
reasoning context. The \emph{candidate cache} $\mathcal{C}_t$ stores chunks
outside the reasoning context, retaining them for later retrieval and
reevaluation. Thus,
$\mathcal{W}_t\cap\mathcal{C}_t=\varnothing$. The cache stores raw retrieved
text rather than model KV states. As shown in
Figure~\ref{fig:qrae_context_flow_framework}(c), \textbf{only whitelisted
retrieved chunks enter the reasoning context}. Chunks in $\mathcal C_t$ remain
available to the evaluator but are excluded from the agent's reasoning input.

\MethodName modifies only the search-snippet portion of the reasoning context.
The instruction, user query, generated reasoning, tool calls, and other tool
outputs are preserved by the underlying agent. After processing retrievals from step $t$,
the next reasoning context therefore replaces the accumulated retrieved chunks
$\mathcal{D}_{\leq t}$ with the current whitelist $\mathcal{W}_t$. The objective
is to retain useful evidence in the reasoning context while keeping uncertain
chunks recoverable in the candidate cache.

\subsection{Online Cache-to-Whitelist Admission}
\label{sec:online-update}

After each search, newly retrieved chunks are deduplicated against the existing
states and merged with previously deferred candidates, yielding the pool
\begin{equation*}
\mathcal U_t=\operatorname{Dedup}(\mathcal C_{t-1}\cup\mathcal D_t)
\setminus\mathcal W_{t-1}.
\end{equation*}
The batch size $K$ equals the agent's configured search-result count per query.
Evaluation is triggered only if $|\mathcal U_t|\geq K$. The scheduler selects
the $K$ candidates with the fewest previous evaluations, prioritizing chunks
that have never been scored. Evaluation occurs at most once per query update, even
if the remaining cache still contains at least $K$ chunks. We write
$\mathcal B_t$ for the selected batch, with $|\mathcal B_t|=K$ at a trigger
and $\mathcal B_t=\varnothing$ otherwise. Candidates outside $\mathcal B_t$
incur no scoring cost at that step. The bound applies to the evaluated batch,
not to the total off-context store in
Figure~\ref{fig:qrae_context_flow_framework}(c).

For each $d\in\mathcal B_t$, the evaluator produces a score
$\operatorname{QRAE}_t(d)$ using the paired prefills and entropy aggregation in
Figure~\ref{fig:qrae_context_flow_framework}(b). Both calls use the same frozen
reasoning LLM as the agent and generate no output tokens. They return the
selected attention responses without modifying model weights or token-level
KV-cache semantics. Section~\ref{sec:qrae-signal} defines this signal.

Let $\mathcal T_t(d)=\{r\leq t:d\in\mathcal B_r\}$ be the steps at which
$d$ has been evaluated. For a nonempty score history, the running mean is
\begin{equation*}
\overline{\operatorname{QRAE}}_t(d)=
\frac{1}{|\mathcal T_t(d)|}
\sum_{r\in\mathcal T_t(d)}\operatorname{QRAE}_r(d).
\end{equation*}
After updating the histories of $\mathcal B_t$, admission ranks all scored
candidates in the entire cache, including those not reevaluated at step $t$.
Let $\mathcal E_t=\{d\in\mathcal U_t:|\mathcal T_t(d)|>0\}$ and let
$\rho\in(0,1]$ denote the retention rate. At a trigger,
\begin{equation*}
\mathcal A_t=\operatorname{Top}_{\lceil\rho K\rceil}
\bigl(\mathcal E_t;\overline{\operatorname{QRAE}}_t\bigr),
\qquad \rho=0.5\ \text{by default}.
\end{equation*}
Here $\operatorname{Top}_j(\mathcal E;s)$ selects the $j$ members of
$\mathcal E$ with the largest scores $s$. All $K$ evaluated candidates have
scores after the update, so $|\mathcal E_t|\geq K$. Thus $K=5$ admits three
chunks and $K=10$ admits five. Admission uses a fixed retention rate without an
absolute score cutoff. Unscored candidates are ineligible. If evaluation is
not triggered, $\mathcal A_t=\varnothing$. The states are then updated as
\begin{equation*}
\mathcal W_t=\mathcal W_{t-1}\cup\mathcal A_t,
\qquad
\mathcal C_t=\mathcal U_t\setminus\mathcal A_t.
\end{equation*}
Admission is persistent: once a chunk enters $\mathcal W_t$, later score
variation does not remove it from the reasoning context. All non-admitted
chunks remain recoverable in $\mathcal C_t$ for later evaluation, while only
$\mathcal W_t$ is included in subsequent reasoning contexts. Algorithm~\ref{alg:qrae-manager}
gives the complete procedure; Appendix~\ref{app:implementation_details}
records the implementation settings.

The manager operates outside the model and requires only access to selected
attention rows during prefill, introducing only lightweight time overhead.
Backend-specific hooks can expose these rows without changing the decoding
path or backend operators.
Appendix~\ref{app:evaluation_cost} analyzes the evaluation overhead and token
savings from deferred admission.

\section{QRAE Signal}
\label{sec:mechanism}
\label{sec:qrae-signal}

\begin{figure}[t]
  \centering
  \includegraphics[width=\textwidth]{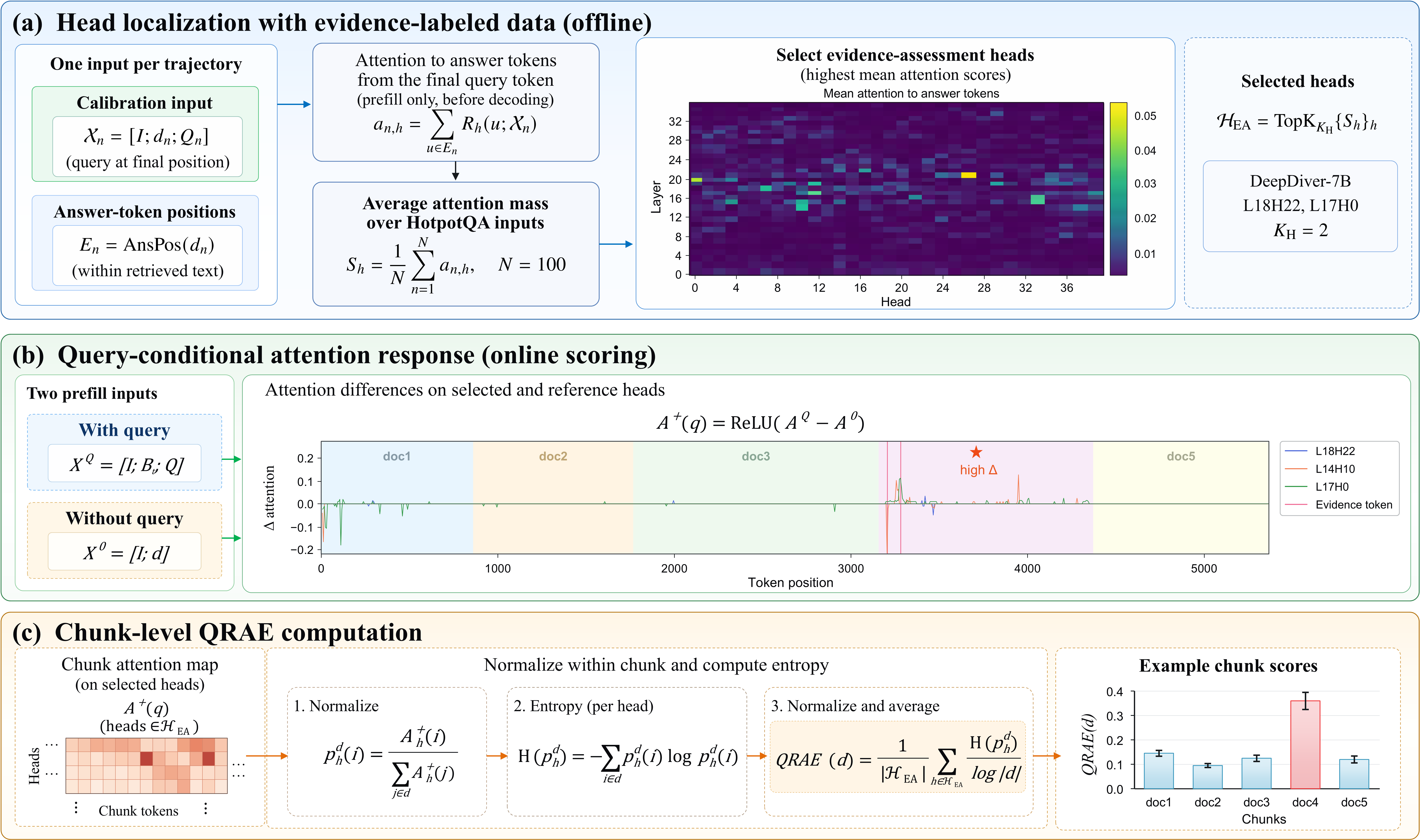}
  \caption{\textbf{Construction of the \SignalName signal.}
  (a) Offline localization selects the model heads with the highest attention
  mass on answer tokens in the input.
  (b) During online scoring, prefills with and without the query yield signed
  attention differences.
  (c) Positive attention differences are first normalized within each chunk.
  Normalized entropy is then computed and averaged across the selected heads.}
  \label{fig:qrae_signal_overview}
\end{figure}

This section defines the model-internal \SignalName signal and details its
scoring procedure for assessing the value of retrieved chunks for subsequent
reasoning. We first localize the model's query-relevant heads offline and then
compute \SignalName during online inference.
Figure~\ref{fig:qrae_signal_overview} summarizes the complete \SignalName
signal construction process.

\subsection{Offline Head Localization}
\label{sec:offline_head_localization}

We localize heads by examining the model's query-conditioned prefill attention
distributions. For each model, we sample 100 HotpotQA trajectories and
select one input from each, requiring that its retrieved text contain the
correct answer tokens. Let
$\mathcal D_{\mathrm{cal}}=\{(Q_n,d_n,E_n)\}_{n=1}^{N}$, where $N=100$ and
$E_n$ marks the answer-token positions in the retrieved text $d_n$. Each
localization input ends with the query:
\begin{equation*}
  \mathcal X_n=[I;d_n;Q_n].
\end{equation*}
At the end of prefill, before decoding, let $R_g(u;\mathcal X_n)$ be the
attention from the final query token to input token $u$ for layer--head pair
$g=(\ell,h)$. Summing over answer tokens and averaging over inputs gives
\begin{equation}
  S_g=\frac{1}{N}\sum_{n=1}^N\sum_{u\in E_n}R_g(u;\mathcal X_n),\qquad
  \mathcal H_{\mathrm{EA}}=\operatorname{TopK}_{K_{\mathrm H}}\{S_g\}_g.
  \label{eq:head_localization}
\end{equation}
The reference answer identifies $E_n$ but is not appended to the input.
The localized head set is a model-level property that generalizes across
datasets and trajectories; it remains fixed throughout online evaluation.
We write $K_{\mathrm H}$ for the selected head count,
distinct from the candidate-batch size $K$. Appendix~\ref{app:implementation_details}
lists the heads and sampling protocol. Appendix~\ref{app:theory} gives a
finite-sample condition for repeatable top-$K_{\mathrm H}$ localization.

\subsection{Online QRAE Score Computation}
\label{sec:qrae_scoring}

During online inference, the \MethodName framework uses the heads localized
offline to compute attention differences against a query-free control, then
computes normalized entropy to obtain the \SignalName signal.

\paragraph{Computing attention differences with a query-free control.}
For a bounded batch $\mathcal B_t$, the \SignalName evaluator constructs inputs with and
without the query,
\begin{equation}
  \mathcal X_t^{Q}=[I;\mathcal B_t;Q], \qquad
  \mathcal X_t^{0}=[I;\mathcal B_t].
  \label{eq:qrae_inputs}
\end{equation}
Here and below, $u$ indexes a token position in a candidate chunk. Each prefill
reads attention from its own final input token: the final query token in
$\mathcal X_t^Q$ and the final candidate-input token in $\mathcal X_t^0$.
The instruction and serialized candidate batch form their common prefix,
which is reused through vLLM prefix caching. Let $h\in\mathcal H_{\mathrm{EA}}$
index a selected layer--head pair. The responses
$\Attention_h(u;\mathcal X_t^Q)$ and $\Attention_h(u;\mathcal X_t^0)$ are
separately normalized to unit mass over the same candidate tokens, as detailed
in Appendix~\ref{app:qrae_computation}. The score uses the positive difference
between these normalized rows:
\begin{equation*}
  \Delta_h(u)=
  \left[
    \Attention_h(u;\mathcal X_t^Q)
    -\Attention_h(u;\mathcal X_t^0)
  \right]_+.
\end{equation*}
The subtraction suppresses response shared by the two prefills, and positive
clipping retains increases in the query-conditioned row. The pairing matches
the instruction and candidate tokens, but the final readout positions differ.
Appendix~\ref{app:theory} bounds the effect of this imperfect matching under
an explicit shared-response decomposition.

The curves in Figure~\ref{fig:qrae_signal_overview}(b) show signed differences
for the selected heads L18H22 and L17H0. The fourth chunk contains the
highlighted evidence tokens and exhibits multiple positive responses.

\paragraph{Computing the QRAE signal with normalized entropy.}
A supporting statement commonly distributes information across several tokens:
an entity, relation, value, qualifier, or temporal constraint may all contribute
to the relevant fact. An incidental match, by contrast, can produce one sharp
response without supporting the complete relation. For each chunk
$d\in\mathcal B_t$, \SignalName measures how broadly the positive response
$\Delta_h$ is distributed over the tokens in $d$ and averages normalized
entropy across the selected heads:
\begin{equation}
  \mathrm{QRAE}_t(d)
  =\frac{1}{|\mathcal H_{\mathrm{EA}}|}
  \sum\nolimits_{h\in\mathcal H_{\mathrm{EA}}}
  \frac{\Entropy\!\left(\{\Delta_h(u)\}_{u\in d}\right)}{\log |d|}.
\label{eq:qrae_entropy}
\end{equation}
Here $\Entropy(\{\Delta_h(u)\}_{u\in d})$ denotes entropy after normalizing the
positive responses within $d$: $p_h^d(u)=\Delta_h(u)/\sum_{v\in d}\Delta_h(v)$
when the denominator is nonzero, and
$\Entropy(p_h^d)=-\sum_{u\in d}p_h^d(u)\log p_h^d(u)$.
These are the normalization and entropy steps in
Figure~\ref{fig:qrae_signal_overview}(c). The score lies in $[0,1]$ and measures
response breadth. Appendix
\ref{app:qrae_computation} gives the complete normalization and zero-mass
convention.

Appendix~\ref{app:robust_entropy_separation} provides a theoretical analysis of
normalized-entropy scoring and its robustness to control mismatch.

\section{Experiments}
\label{sec:experiments}

We organize the experimental evaluation in the following order to validate
the effectiveness of \MethodName.
\textbf{RQ1 (DeepSearch agent benefits):} Can \MethodName reduce latency and
token costs while maintaining task performance?
\textbf{RQ2 (signal validity):} Can the \SignalName score consistently and
reliably identify useful evidence across datasets and models?
\textbf{RQ3 (component contributions):} We evaluate whether query-free
subtraction, entropy scoring, and context management provide benefits.
We first present the benefits for DeepSearch agents, then analyze the
\SignalName signal and the individual components.

\subsection{Experimental Setup}
\label{sec:setup}

\paragraph{Benchmarks.}
We use established DeepSearch benchmarks covering diverse task types and
difficulty levels: BrowseComp-Plus~\citep{chen2025browsecompplus} for local
retrieval, and XBench-DeepSearch~\citep{chen2025xbench},
FRAMES~\citep{krishna2024frames}, and SimpleQA for online retrieval.
We use HotpotQA solely to localize \EAHs, then evaluate the effectiveness and
efficiency gains of \MethodName on the remaining benchmarks with these heads
fixed to assess model-level generalization.

\paragraph{Models.}
We use DeepDiver-7B, a model designed for DeepSearch, as the primary model for
evaluating \MethodName. To assess cross-model generalization, we also localize
heads and evaluate \SignalName scoring in the DeepSearch models LiteResearcher-4B,
WebSeer-14B, and OpenResearcher-30B~\citep{li2026literesearcher,he2026webseer,li2026openresearcher},
as well as OpenPangu-2.0-Flash-92B with nearly 100B parameters~\citep{openpangu2026flash}.

\paragraph{Baselines.}
We use \textbf{DeepSearch} as the uncompressed baseline. Under matched inference
backends, we compare \MethodName with this baseline and the compression methods
LLMLingua-2, ACON, Rerank, QRHead, and SideQuest~\citep{pan2024llmlingua2,kang2026acon,
sharifymoghaddam2026rerank,zhang2025qrhead,kariyappa2026sidequest}.
These baselines cover different compression strategies: LLMLingua-2 removes
tokens using a learned general-purpose compression policy; ACON rewrites agent
history through LLM summarization; Rerank reranks retrieved results;
QRHead uses selected attention heads for text selection;
and SideQuest combines context management with inference-level KV-cache
optimization.

\paragraph{Metrics.}
We evaluate all methods on all trajectories of each dataset. We report the
overall task success rate and per-trajectory means of end-to-end latency and
cumulative token consumption. Unless otherwise specified, tokens include all
model inputs, outputs, and \SignalName evaluator inputs; latency includes
inference, retrieval, webpage processing, and evaluation time. All methods are
run serially to ensure a fair comparison. Appendix~\ref{app:experimental_details}
provides the detailed experimental settings and evaluation protocol.

\subsection{QRAE Effectiveness in DeepSearch Agents}
\label{sec:end_to_end}

We first evaluate \MethodName with DeepDiver-7B against the full set of
baselines across all four benchmarks. Table~\ref{tab:end_to_end} reports
task success, end-to-end latency, and total processed tokens to jointly
assess task effectiveness and computational efficiency.

\begin{table}[ht]
\centering
\caption{\MethodName effectiveness in DeepSearch agents with DeepDiver-7B
across four benchmarks. Results report task success, end-to-end time, and
total processed tokens from one representative run per benchmark. Bold and
underlined values denote the best and second-best results, respectively.}
\label{tab:end_to_end}
\normalsize
\setlength{\tabcolsep}{2pt}
\renewcommand{\arraystretch}{1.02}
\begin{tabular*}{\textwidth}{@{\extracolsep{\fill}}llccccccc@{}}
\toprule
\multicolumn{1}{c}{\multirow{2}{*}{Dataset}}
& \multicolumn{1}{c}{\multirow{2}{*}{Metric}}
& DeepSearch & Lingua-2 & ACON & Rerank & QRHead & SideQuest & \textbf{\MethodName} \\
\cmidrule(lr){3-3}\cmidrule(lr){4-8}\cmidrule(lr){9-9}
& & \multicolumn{1}{c}{No comp.}
& \multicolumn{5}{c}{Compression baselines}
& \multicolumn{1}{c}{\textbf{Ours}} \\
\midrule
\multirow{3}{*}{\shortstack[l]{BrowseComp-\\Plus}}
& Succ. (\%) $\uparrow$ & 13.8 & 14.3 & 12.1 & \underline{14.7} & 11.6 & 10.6 & \textbf{17.2} \\
& Time (s) $\downarrow$ & 29.5 & 28.8 & 50.0 & 35.6 & 45.1 & \underline{24.3} & \textbf{20.5} \\
& Tokens $\downarrow$ & 177k & 171k & 98k & 148k & 149k & \underline{66k} & \textbf{54k} \\
\cmidrule(lr){1-9}
\multirow{3}{*}{\shortstack[l]{XBench-\\DeepSearch}}
& Succ. (\%) $\uparrow$ & 38.0 & \underline{43.0} & 40.0 & 41.0 & 35.0 & 36.0 & \textbf{44.0} \\
& Time (s) $\downarrow$ & 276.0 & 245.3 & 324.8 & 213.6 & 357.5 & \underline{167.2} & \textbf{150.4} \\
& Tokens $\downarrow$ & 319k & 288k & 146k & 227k & 303k & \underline{144k} & \textbf{117k} \\
\cmidrule(lr){1-9}
\multirow{3}{*}{FRAMES}
& Succ. (\%) $\uparrow$ & 40.9 & 41.2 & 39.5 & \textbf{43.1} & 37.3 & 38.9 & \underline{42.8} \\
& Time (s) $\downarrow$ & 158.7 & 155.2 & 184.3 & 114.4 & 203.7 & \underline{105.8} & \textbf{90.7} \\
& Tokens $\downarrow$ & 140k & 137k & 105k & 102k & 124k & \underline{99k} & \textbf{94k} \\
\cmidrule(lr){1-9}
\multirow{3}{*}{SimpleQA}
& Succ. (\%) $\uparrow$ & 79.6 & 79.9 & \underline{80.4} & 77.6 & 78.2 & 75.9 & \textbf{82.2} \\
& Time (s) $\downarrow$ & 145.6 & 141.3 & 132.8 & 120.5 & 194.5 & \underline{106.5} & \textbf{98.6} \\
& Tokens $\downarrow$ & 74k & 73k & 52k & 60k & 65k & \textbf{38k} & \underline{45k} \\
\bottomrule
\end{tabular*}
\end{table}

Relative to the uncompressed DeepSearch agent, \MethodName improves success
by up to 6.0 percentage points. In settings with larger gains, processed tokens
and latency fall by roughly 60\% and 30\%, respectively. Among compression
methods, \MethodName offers the best overall balance between task effectiveness
and computational efficiency. Appendix~\ref{app:run_stability} reports
mean $\pm$ STD over five runs with different random seeds. Higher mean success
and lower mean token consumption and latency across all four benchmarks
support the stability of these gains.

To assess transfer across model architectures, we extend the full baseline
comparison to OpenResearcher-30B with its hybrid Mamba--Transformer
architecture~\citep{li2026openresearcher} and OpenPangu-2.0-Flash-92B with
hybrid sparse attention~\citep{openpangu2026flash}. We use BrowseComp-Plus,
the most challenging benchmark in our evaluation, whose fixed local corpus
enables reproducible retrieval. Table~\ref{tab:cross_model_end_to_end} reports
task success, end-to-end latency, and cumulative processed tokens. The results
show that \MethodName remains effective in architectures with linear-time
sequence modeling or sparse attention.

\begin{table}[H]
\centering
\caption{Cross-model evaluation of \MethodName on BrowseComp-Plus, reporting
task success, end-to-end time, and total processed tokens. Token counts for
OpenPangu include inputs only. Bold and underlined values denote the best
and second-best results, respectively.}
\label{tab:cross_model_end_to_end}
\normalsize
\setlength{\tabcolsep}{4pt}
\renewcommand{\arraystretch}{1.12}
\begin{tabular*}{\textwidth}{@{\extracolsep{\fill}}lcccccc@{}}
\toprule
\multirow{2}{*}{Method}
& \multicolumn{3}{c}{OpenResearcher-30B}
& \multicolumn{3}{c}{OpenPangu-2.0-Flash-92B} \\
\cmidrule(lr){2-4}\cmidrule(lr){5-7}
& Success (\%) $\uparrow$ & Time (s) $\downarrow$ & Tokens $\downarrow$
& Success (\%) $\uparrow$ & Time (s) $\downarrow$ & Tokens $\downarrow$ \\
\midrule
DeepSearch                  & \textbf{53.2} & 151.0 & 1981k & \underline{54.9} & 308.6 & 1077k \\
LLMLingua-2           & 52.1 & 154.7 & 2037k & 54.2 & 304.5 & 1057k \\
ACON                  & 49.4 & 188.1 & \textbf{842k} & 50.1 & 429.8 & 879k \\
Rerank                & 50.7 & 162.0 & 2157k & 54.3 & 344.7 & 970k \\
QRHead                & 51.1 & 189.3 & 1771k & 53.8 & 393.8 & 982k \\
SideQuest             & 47.6 & \underline{144.6} & 1605k & 51.3 & \underline{284.5} & \underline{801k} \\
\textbf{\MethodName}   & \underline{53.0} & \textbf{134.7} & \underline{1453k} & \textbf{55.3} & \textbf{255.4} & \textbf{660k} \\
\bottomrule
\end{tabular*}
\end{table}

\subsection{QRAE Signal Validity}
\label{sec:signal_validation}

\begin{figure}[H]
  \centering
  \captionsetup[subfigure]{skip=2pt}
  \begin{subfigure}[b]{0.33\textwidth}
    \centering
    \includegraphics[width=\linewidth]{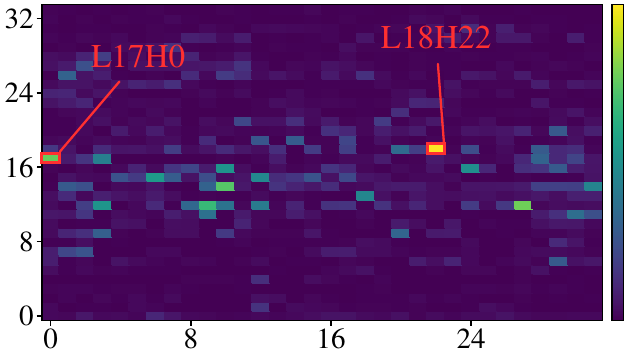}
    \caption{HotpotQA}
  \end{subfigure}\hfill
  \begin{subfigure}[b]{0.33\textwidth}
    \centering
    \includegraphics[width=\linewidth]{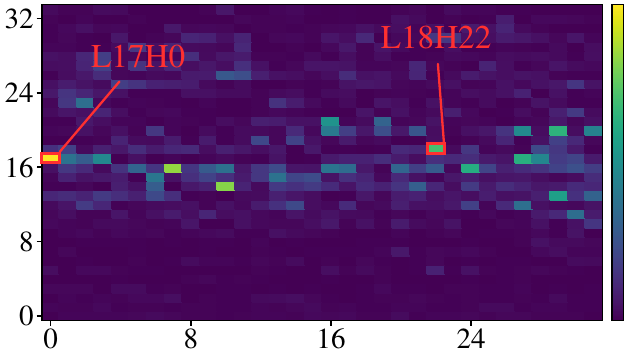}
    \caption{BrowseComp-Plus}
  \end{subfigure}\hfill
  \begin{subfigure}[b]{0.33\textwidth}
    \centering
    \includegraphics[width=\linewidth]{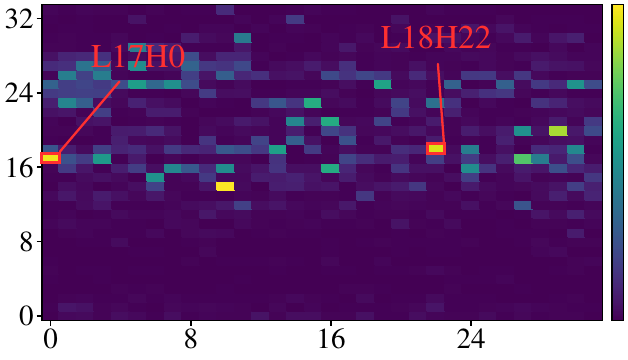}
    \caption{XBench-DeepSearch}
  \end{subfigure}
  \caption{Global layer--head localization scores for DeepDiver-7B, computed
  from attention to answer-bearing spans. Heads are selected on HotpotQA (a) and validated on
  BrowseComp-Plus (b) and XBench-DeepSearch (c), neither of which is used for
  head selection.}
  \label{fig:head_localization}
\end{figure}

We localize \EAHs in DeepDiver-7B using only HotpotQA and keep the selected
head set fixed for subsequent evaluation. Figure~\ref{fig:head_localization}
presents the head-localization results: L18H22 and L17H0 consistently receive
the highest localization scores across all evaluated datasets, indicating
that the localized heads are model-level and remain unchanged across datasets.

Appendix~\ref{app:cross_model_heads} provides head-localization results for
LiteResearcher-4B, WebSeer-14B, OpenResearcher-30B, and
OpenPangu-2.0-Flash-92B. The agent-level benefits of \MethodName in
OpenResearcher-30B and OpenPangu-2.0-Flash-92B are evaluated in
Section~\ref{sec:end_to_end}.

We compare \SignalName scoring across all five models using the same
BrowseComp-Plus candidate sets of 5 and 20 chunks, with each model's
localized heads fixed. AUC denotes AUROC and measures how well the scores
distinguish evidence-bearing chunks from other candidates overall.
Recall@1 tests the extreme retention setting in which only the top-ranked
chunk is kept. Recall@3 for 5 chunks and Recall@10 for 20 chunks assess
evidence recall when retaining multiple chunks, more closely reflecting
context filtering during agent execution. Recall@$k$ is the fraction of
annotated evidence chunks in the candidate set retained among the top $k$.

% Candidate-set construction and metric aggregation are specified in
% Appendix~\ref{app:benchmark_protocol}.
% Measured values for all models supplied by the author; all metrics are
% expressed as percentages.
\begin{table}[H]
\centering
\caption{Cross-model \SignalName scoring on BrowseComp-Plus with 5- and
20-chunk candidate sets. AUC denotes AUROC; all metrics are percentages
and higher is better.}
\label{tab:cross_model_ranking}
\normalsize
\setlength{\tabcolsep}{2pt}
\renewcommand{\arraystretch}{1.12}
\begin{tabular*}{\textwidth}{@{\extracolsep{\fill}}lcccccc@{}}
\toprule
\multirow{2}{*}{Model}
& \multicolumn{3}{c}{5 chunks}
& \multicolumn{3}{c}{20 chunks} \\
\cmidrule(lr){2-4}\cmidrule(lr){5-7}
& AUC & Recall@1 & Recall@3
& AUC & Recall@1 & Recall@10 \\
\midrule
LiteResearcher-4B  & 76.6 & 48.3\% & 76.6\% & 81.8 & 35.2\% & 83.2\% \\
DeepDiver-7B       & 83.2 & 54.6\% & 83.2\% & 84.4 & 44.6\% & 88.7\% \\
WebSeer-14B        & 84.1 & 53.9\% & 84.5\% & 85.1 & 47.8\% & 89.2\% \\
OpenResearcher-30B & 90.3 & 74.1\% & 90.5\% & 89.6 & 57.4\% & 93.0\% \\
OpenPangu-2.0-Flash-92B & 92.5 & 83.8\% & 93.4\% & 92.2 & 66.2\% & 96.8\% \\
\bottomrule
\end{tabular*}
\end{table}

Table~\ref{tab:cross_model_ranking} shows that \SignalName exhibits nontrivial
evidence-scoring ability in all five models, with AUC exceeding 75\% in both
candidate-set settings. Larger models generally achieve higher AUC and recall,
indicating stronger evidence discrimination and retention. Attention
interventions further support a functional role for the localized
heads in subsequent reasoning (Appendix~\ref{app:expanded_intervention}).

\subsection{Ablation Study}
\label{sec:ablations}

We examine the contributions of \MethodName's components through three variants.
All variants use DeepDiver-7B with the same localized head set.

\begin{itemize}[leftmargin=*,nosep]
\item \textbf{w/o query-free subtraction.} Remove the query-free prefill and compute normalized entropy directly
from query-conditioned attention, retaining within-chunk normalization and the
same context-management policy.
\item \textbf{Attention sum.} Retain query-free subtraction and context management, and replace entropy
with summed positive attention responses averaged over the selected heads.
\item \textbf{Current-search filtering.} Apply the full \SignalName score only to current search results, retaining
selected chunks in the history and discarding the rest, without a persistent
whitelist or a cache for later reevaluation.
\end{itemize}

We evaluate scoring quality and agent performance. The former tests evidence
ranking with fixed candidate sets and a fixed number of retained chunks;
the latter measures task success and the actual costs
of the resulting agent trajectories.

\paragraph{Scoring quality.}
Table~\ref{tab:scoring_ablations} compares the full \SignalName score with
the two scoring variants on BrowseComp-Plus, using the same 5- and 20-chunk
candidate sets and six metrics as in Section~\ref{sec:signal_validation}.
Current-search filtering is evaluated only at the agent level because it uses
the unchanged \SignalName score and modifies cross-step context management.

Both scoring ablations consistently reduce AUC and every recall metric for
both candidate-set sizes. The drop without query-free subtraction supports
the utility of suppressing responses shared with the query-free control. With this
subtraction held fixed, entropy outperforms attention summation on every
metric, indicating that response breadth better identifies useful evidence
than total attention mass.

% Scoring ablation values supplied by the author.
\begin{table}[H]
\centering
\caption{Scoring ablations on BrowseComp-Plus with DeepDiver-7B.
Candidate sets and retained-chunk budgets are shared across variants.
AUC denotes AUROC; all metrics are percentages and higher is better.}
\label{tab:scoring_ablations}
\normalsize
\setlength{\tabcolsep}{2pt}
\renewcommand{\arraystretch}{1.12}
\begin{tabular*}{\textwidth}{@{\extracolsep{\fill}}lcccccc@{}}
\toprule
\multirow{2}{*}{Variant}
& \multicolumn{3}{c}{5 chunks}
& \multicolumn{3}{c}{20 chunks} \\
\cmidrule(lr){2-4}\cmidrule(lr){5-7}
& AUC & Recall@1 & Recall@3
& AUC & Recall@1 & Recall@10 \\
\midrule
\textbf{\SignalName\ (full)} & \textbf{83.2} & \textbf{54.6\%} & \textbf{83.2\%} & \textbf{84.4} & \textbf{44.6\%} & \textbf{88.7\%} \\
w/o query-free subtraction & 79.6 & 48.6\% & 80.9\% & 78.8 & 34.2\% & 84.5\% \\
Attention sum & 75.0 & 44.7\% & 79.5\% & 81.3 & 32.6\% & 80.0\% \\
\bottomrule
\end{tabular*}
\end{table}

\paragraph{End-to-end agent performance.}
Table~\ref{tab:component_ablations} compares the full method with all three
variants on BrowseComp-Plus and XBench-DeepSearch, using the same retrieval
settings and all evaluation trajectories. Success, latency, and processed
tokens capture the task-level effects of both scoring and context management,
including the evaluator overhead required by each variant.

% Values supplied by the author as measured ablation results.
% TODO: Verify that all context-management variants use the reported retention-rate settings.
\begin{table}[H]
\centering
\caption{Agent-level component ablations on BrowseComp-Plus and XBench-DeepSearch.
Success is in percent, time in seconds, and tokens in thousands.
End-to-end costs include the evaluator calls required by each variant.}
\label{tab:component_ablations}
\normalsize
\setlength{\tabcolsep}{3pt}
\renewcommand{\arraystretch}{1.12}
\begin{tabular*}{\textwidth}{@{\extracolsep{\fill}}lcccccc@{}}
\toprule
\multirow{2}{*}{Variant}
& \multicolumn{3}{c}{BrowseComp-Plus}
& \multicolumn{3}{c}{XBench-DeepSearch} \\
\cmidrule(lr){2-4}\cmidrule(lr){5-7}
& Succ. (\%) $\uparrow$ & Time (s) $\downarrow$ & Tokens $\downarrow$
& Succ. (\%) $\uparrow$ & Time (s) $\downarrow$ & Tokens $\downarrow$ \\
\midrule
\textbf{\MethodName\ (full)}
& \textbf{17.2} & \textbf{20.5} & \textbf{54}
& \textbf{44.0} & \textbf{150.4} & \textbf{117} \\
w/o query-free subtraction & 15.8 & 24.1 & 73 & 42.0 & 192.8 & 186 \\
Attention sum & 14.6 & 27.3 & 92 & 42.0 & 181.6 & 177 \\
Current-search filtering & 14.9 & 30.2 & 129 & 41.0 & 232.3 & 225 \\
\bottomrule
\end{tabular*}
\end{table}

Every ablation lowers success while increasing token consumption and latency
on both datasets, supporting the contribution of all three components.
For the two scoring variants, these losses align with their weaker evidence
ranking in Table~\ref{tab:scoring_ablations}. Current-search filtering retains
the full scorer but loses chunk reevaluation over the evolving trajectory.
When one search returns multiple query-relevant chunks, immediate filtering
can discard evidence needed later. Recovering that evidence may require
additional reasoning steps and tool calls, providing a plausible explanation
for the higher end-to-end costs despite filtering each search.

Appendix~\ref{app:scoring_robustness} examines the sensitivity of \SignalName
AUC and recall to input-chunk permutations within fixed candidate sets.

\section{Conclusion}
\label{sec:conclusion}

We introduced \MethodName, an online context compression method that uses
query-relevant attention entropy to select retrieved chunks for DeepSearch
agents. Persistent admission keeps useful chunks in subsequent contexts,
while cached candidates remain available for reevaluation.

Averaged equally across six model--benchmark settings, \MethodName
improves task success by 2.35 percentage points and reduces reported token
usage and end-to-end latency by approximately 45\% and 30\%, respectively,
relative to uncompressed DeepSearch.
Both the signal and method generalize across the evaluated datasets and model
architectures, including hybrid Mamba--Transformer and sparse-attention models.

Prefill-only scoring adds little overhead and integrates with existing
inference backends through hooks, without modifying decoding or backend operators.
Future work will clarify how the localized heads route information into
subsequent reasoning and extend \MethodName to tasks whose evidence cannot
be localized to retrieved text.

\clearpage
\subsection*{AI use statement}
We used generative AI tools solely for language polishing to improve the clarity and readability of
the manuscript. No generative AI tools were used for research ideation, theoretical
proofs, data processing, or experimental design.
The authors take full responsibility for the content of this paper.

\subsection*{Ethics statement}
This work aims to improve the inference efficiency of DeepSearch agents.
We do not identify specific ethical concerns associated with the proposed method.

\subsection*{Reproducibility statement}
Sections~\ref{sec:framework} and~\ref{sec:mechanism} specify the framework and
scoring rule. Appendix~\ref{app:method_details} provides pseudocode, normalization
conventions, selected heads, and implementation details.
Appendix~\ref{app:experimental_details} documents datasets, model checkpoints,
agent and retrieval configurations, baseline adaptations, hardware, cost
accounting, and random seeds. Appendix~\ref{app:additional_results} reports
robustness checks, repeated-run results, and cost breakdowns, while
Appendix~\ref{app:theory} states the assumptions and proofs for the theoretical
analysis.

\bibliography{token}
\bibliographystyle{iclr2027_conference}

\clearpage
\appendix

\section{Related Work}
\label{sec:related_work}
\label{app:related_work}

\subsection{Agent Context Management}

Long-horizon agents accumulate reasoning, tool calls, and observations that may
be reprocessed across many steps. ACON condenses observations and histories with
an optimized compressor, while Context-Folding and AgentFold learn to replace
completed trajectory regions with compact folds~\citep{kang2026acon,
sun2025contextfolding,ye2025agentfold}. RE-TRAC converts completed search
trajectories into structured states that guide later exploration
\citep{zhu2026retrac}. PACE reconstructs working context from multigranularity
memory using predicted next-step relevance, whereas ARC monitors and revises
context through reflection~\citep{wei2026pace,yao2026arc}.

These methods establish context management as an active part of the agent, but
their primary representations are condensations, folds, structured states, or
revised histories. \MethodName instead makes forwarding decisions over intact
retrieved chunks. Whitelisted chunks persist in the reasoning context; deferred
chunks remain outside it in the candidate cache as raw text and can be retrieved
and rescored when the trajectory changes. The policy therefore avoids a
generative rewriting step while leaving reasoning, tool interactions, and
document text unchanged.

\subsection{Prompt Evidence Selection}

Rerank Before You Reason selects retrieved passages before search-time
reasoning, while LLMLingua-2 predicts token importance for single-prompt
compression~\citep{sharifymoghaddam2026rerank,pan2024llmlingua2}. Attention-based
methods derive selection signals from the model itself. EHPC identifies
evaluator heads in early prefill layers and retains important prompt tokens;
AttnComp likewise compresses long inputs using attention-derived importance
\citep{fei2025ehpc,zhao2025attncomp}. Retrieval heads are causally implicated in
long-context factual recall, and QRHead aggregates query-focused attention for
retrieval and reranking~\citep{wu2025retrievalhead,zhang2025qrhead}.

The \SignalName evaluator differs in both measurement and use. It localizes
heads from final-query-token attention to answer tokens before decoding,
then uses an online query-free control and scores
the breadth of the positive response through normalized within-chunk entropy.
Rather than deleting tokens for one call or reranking one retrieval set, the
score drives persistent admission and recoverable deferral across agent steps.
The intervention experiments further test whether the selected heads are
functionally connected to downstream reasoning.

\subsection{Inference-State Compression}

KV-cache methods reduce memory or attention work by retaining selected token
states. H2O protects recent and high-attention tokens, SnapKV estimates important
positions from an observation window, and DuoAttention allocates full caches to
retrieval heads~\citep{zhang2023h2o,li2024snapkv,xiao2024duoattention}. For
agent trajectories, SideQuest asks the reasoning model to manage KV state,
IntentKV performs cross-turn intent-aware pruning, and AgentInfer co-designs
context management with the inference system~\citep{kariyappa2026sidequest,
li2026intentkv,lin2025agentinfer}. These approaches change which cached states
participate in inference. \MethodName instead changes which retrieved text
enters future prefills while preserving token-level cache semantics, so the two
families are complementary.

Across these lines, \MethodName focuses on the intersection of intact chunk
granularity, model-internal prefill scoring, persistent cross-step admission,
recoverable deferral, and bounded evaluation. This positioning motivates the
heterogeneous baselines in Section~\ref{sec:setup}.

\section{Method Details}
\label{app:method_details}

\subsection{Online Context-Management Procedure}
\label{app:online_procedure}

\begin{algorithm}[H]
\caption{Recoverable context admission with \MethodName}
\label{alg:qrae}
\label{alg:qrae-manager}
\begin{algorithmic}[1]
\Require Query $Q$, selected heads $\mathcal H_{\mathrm{EA}}$, search-result count
$K$, retention rate $\rho=0.5$
\State $\mathcal C\gets\emptyset$, $\mathcal W\gets\emptyset$
\For{each search at agent step $t$}
    \State Retrieve $\mathcal D_t$; add chunks not already in $\mathcal C\cup\mathcal W$ to $\mathcal C$
    \State Initialize an empty score history for each new chunk
    \If{$|\mathcal C|\geq K$} \Comment{At most one evaluation per search}
        \State Select the $K$ chunks with the fewest recorded scores as $\mathcal B\subseteq\mathcal C$
        \State Score $\mathcal B$ with query/control prefills and normalized entropy
        \State Append each new score to its chunk history; update its running mean
        \State $\mathcal E\gets\{d\in\mathcal C:d\text{ has at least one recorded score}\}$
        \State $\mathcal A\gets\operatorname{Top}_{\lceil\rho K\rceil}(\mathcal E;\overline{\operatorname{QRAE}})$
        \State $\mathcal W\gets\mathcal W\cup\mathcal A$, $\mathcal C\gets\mathcal C\setminus\mathcal A$
    \EndIf
    \State Run the next reasoning step with reasoning context
    $\mathcal P_{t+1}[\mathcal D_{\leq t}\leftarrow\mathcal W]$
\EndFor
\end{algorithmic}
\end{algorithm}

\subsection{Full QRAE Computation}
\label{app:qrae_computation}

Given the query and control inputs in Equation~\ref{eq:qrae_inputs}, let
$q_t^z=|\mathcal X_t^z|$ be the final input-token index, using one-based
indices. Let $R_h(u;\mathcal X_t^z)$ denote the attention weight from row
$q_t^z$ to candidate token $u$ for selected head $h$ and input $z\in\{Q,0\}$.
One attention row is read from each prefill, with no aggregation over query
positions and no decoding. The \SignalName evaluator first normalizes each response over all candidate tokens
in the evaluation batch:
\begin{equation*}
  \Attention_h(u;\mathcal X_t^z)=
  \frac{R_h(u;\mathcal X_t^z)}
  {\sum\nolimits_{d'\in\mathcal B_t}\sum\nolimits_{v\in d'}R_h(v;\mathcal X_t^z)},
  \qquad z\in\{Q,0\}.
\end{equation*}
Consequently, each normalized row has total candidate-token mass one
(up to numerical precision) before subtraction. This matches the total mass
being compared; it does not make the two final-token readout positions
identical. The evaluator then takes the positive difference and
normalizes it within each chunk:
\begin{align*}
  \Delta_h(u)
  &=\left[
    \Attention_h(u;\mathcal X_t^Q)
    -\Attention_h(u;\mathcal X_t^0)
  \right]_+,
  \\
  m_h(d)
  &=\sum\nolimits_{v\in d}\Delta_h(v),\\
  p_h^d(u)
  &=\frac{\Delta_h(u)}{m_h(d)},
  \qquad u\in d,\quad m_h(d)>0.
\end{align*}
The per-head contribution and final chunk score are
\begin{align*}
  e_h(d)
  &=
  \begin{cases}
    \Entropy(p_h^d)/\log|d|, & m_h(d)>0\ \text{and}\ |d|>1,\\
    0, & \text{otherwise},
  \end{cases}\\
  \mathrm{QRAE}_t(d)
  &=\frac{1}{|\mathcal H_{\mathrm{EA}}|}
  \sum\nolimits_{h\in\mathcal H_{\mathrm{EA}}}e_h(d).
\end{align*}
Here $\Entropy(p)=-\sum_u p(u)\log p(u)$ with $0\log0=0$. Thus a head assigning
no positive difference mass to a chunk contributes zero, while every
nonzero contribution lies in $[0,1]$.

\subsection{Implementation Settings and Head Localization}
\label{app:implementation_details}

\paragraph{Offline inputs and heatmaps.}
For each model, localization uses 100 randomly sampled HotpotQA trajectories,
with one input selected per trajectory. Each selected input must contain
correct-answer tokens in its retrieved text. The input is $[I;d;Q]$ and ends
with the query; the reference answer is used only to identify the answer-token
positions within $d$. At the end of prefill, before any answer is generated,
we read the attention row of the final query token. For each head, we sum its
attention over these answer-token positions and average over the 100 inputs,
as in Equation~\ref{eq:head_localization}. The resulting layer--head matrix is
the localization heatmap. There is no offline subtraction or negative-chunk
pairing. This prefill-only criterion measures attention to evidence already
in the input, rather than attention while generating or copying an answer
during decoding, distinguishing it from a decoding-time retrieval-head criterion.

The heads annotated in Figures~\ref{fig:head_localization}
and~\ref{fig:additional_model_heads} are
the selected sets, listed in Table~\ref{tab:selected_heads}. For every model,
the set is chosen on HotpotQA and held fixed online. The BrowseComp-Plus and
XBench-DeepSearch heatmaps use the same measurement on 100 randomly sampled,
answer-containing inputs per dataset, one per trajectory. These heatmaps
evaluate cross-dataset generalization and do not alter the selected sets.

\begin{table}[H]
\centering
\caption{Fixed evidence-assessment heads. Layer and head indices are zero-based.}
\label{tab:selected_heads}
\normalsize
\begin{tabular}{@{}lcl@{}}
\toprule
Model & $K_{\mathrm H}$ & Selected layer--head pairs \\
\midrule
LiteResearcher-4B & 2 & L20H15, L18H15 \\
DeepDiver-7B & 2 & L18H22, L17H0 \\
WebSeer-14B & 2 & L36H21, L47H33 \\
OpenResearcher-30B & 3 & L26H9, L26H25, L19H18 \\
OpenPangu-2.0-Flash-92B & 5 & L45H36, L39H24, L39H30, L39H32, L36H41 \\
\bottomrule
\end{tabular}
\end{table}

\paragraph{Online readout and prefix reuse.}
Online inputs are $[I;\mathcal B_t;Q]$ and $[I;\mathcal B_t]$, with the same
instruction, candidate ordering, and candidate serialization. Each call reads
its own last input token's attention before decoding. The query-free call ends
at the candidate input, and the query-conditioned call appends $Q$ to that
same prefix. vLLM prefix caching is enabled to reuse the common prefix's KV
states. This serving cache is separate from the external candidate cache,
which stores raw chunk text and score histories. Both evaluator inputs remain
included in the processed-token accounting, while prefix reuse affects the
computation actually performed. Each attention row is normalized over the same
candidate-token set before positive subtraction, followed by within-chunk
normalization and entropy as specified in Appendix~\ref{app:qrae_computation}.

\paragraph{Search-aligned chunking and scheduling.}
The evaluator treats each search-returned snippet as one chunk, preserving the
search configuration's splitting rules and length limits without additional
segmentation. The batch size $K$ equals the configured number of search
results per query. Each query's deduplicated results update the shared cache.
If at least $K$ candidates are available, exactly one batch of $K$ is scored;
otherwise scoring is deferred. Priority is determined by the number of prior
scores, in ascending order. Thus ``low confidence'' in scheduling means fewer
measurements, not a lower mean relevance score. Newly retrieved, unscored
chunks have priority; the number introduced by one search is at most $K$.

\paragraph{Retention and persistent admission.}
After the batch is scored, each evaluated chunk's history and arithmetic mean
are updated. All scored chunks in the entire candidate cache are then ranked
by their historical means; candidates not reevaluated in this batch retain
their previous means. The top $\lceil\rho K\rceil$ enter the whitelist, with
$\rho=0.5$ by default: three chunks for $K=5$ or five for $K=10$. The quota
depends on the batch size, not on the total cache size. Unscored chunks are
excluded from admission. Admitted chunks are removed from the candidate cache
and persist in subsequent reasoning contexts. The others remain available for
future evaluation. The manager does not score a second batch within the same
query update, even when the remaining cache still has at least $K$ members.

\begin{table}[H]
\centering
\caption{Implementation settings shared with the agent or fixed by \MethodName.}
\label{tab:implementation_settings}
\normalsize
\begin{tabular}{@{}p{0.27\textwidth}p{0.68\textwidth}@{}}
\toprule
Setting & Value or rule \\
\midrule
Calibration & 100 HotpotQA trajectories; one answer-containing input each \\
Head set & Fixed per model; Table~\ref{tab:selected_heads} \\
Batch size $K$ & Agent's configured search-result count per query \\
Chunking & Agent's search splitting rules and length limits \\
Trigger & Candidate count at least $K$; at most once per query update \\
Batch selection & $K$ candidates with the fewest prior scores \\
Retention rate $\rho$ & $0.5$; admit $\lceil\rho K\rceil$ chunks per trigger \\
Admission ranking & Historical mean over all scored candidates in the cache \\
Online evaluator & Final-token prefill attention; vLLM prefix caching enabled \\
\bottomrule
\end{tabular}
\end{table}
Experiment-specific search and chunking parameters, model revisions, and
serving settings are collected in Appendix~\ref{app:experimental_details}.
% TODO(author): Supply per-benchmark search-result counts, chunk lengths and
% splitting options, exact evaluator/chat templates, calibration sample IDs,
% model/tokenizer/backend revisions, and tie-breaking rules from the code/config.
% TODO(author): Reconcile the archived heatmap export metadata with the confirmed
% 100-input localization protocol before releasing the reproduction package.

\section{Experimental Settings and Evaluation Protocol}
\label{app:experimental_details}

This appendix specifies the evaluation tasks, model and agent settings,
retrieval and compression configurations, serving environment, and repeated-run
protocol. Appendix~\ref{app:method_details} defines the scoring and context-management
procedures, including the fixed head sets in Table~\ref{tab:selected_heads}.

\subsection{Benchmarks and Evaluation}
\label{app:benchmark_protocol}

\paragraph{Benchmarks and coverage.}
We evaluate end-to-end performance on the full evaluation sets of
BrowseComp-Plus, XBench-DeepSearch, FRAMES, and SimpleQA, without filtering
questions. We use the XBench-DeepSearch-2510 release, the FRAMES test split,
and the original SimpleQA test set. BrowseComp-Plus uses its fixed local
corpus.\footnote{Official releases:
\href{https://huggingface.co/datasets/Tevatron/browsecomp-plus}{BrowseComp-Plus},
\href{https://huggingface.co/datasets/xbench/DeepSearch-2510}{XBench-DeepSearch-2510},
\href{https://huggingface.co/datasets/google/frames-benchmark}{FRAMES}, and
\href{https://github.com/openai/simple-evals/blob/main/simpleqa_eval.py}{SimpleQA}.}
HotpotQA is used for offline head localization and controlled evidence
analysis. Table~\ref{tab:benchmark_protocol} summarizes the datasets and
their roles.

\begin{table}[H]
\centering
\caption{Datasets and evaluation coverage. All questions in the four
end-to-end benchmarks are evaluated.}
\label{tab:benchmark_protocol}
\normalsize
\setlength{\tabcolsep}{4pt}
\renewcommand{\arraystretch}{1.15}
\begin{tabular}{@{}>{\raggedright\arraybackslash}p{0.26\textwidth}
                  >{\raggedright\arraybackslash}p{0.41\textwidth}
                  >{\raggedright\arraybackslash}p{0.27\textwidth}@{}}
\toprule
Benchmark & Role & Evaluation set \\
\midrule
BrowseComp-Plus & End-to-end and evidence ranking & 830 questions \\
XBench-DeepSearch & End-to-end and head transfer & 2510 release; 100 questions \\
FRAMES & End-to-end multi-hop QA & Test split; 824 questions \\
SimpleQA & End-to-end factual QA & Original test set; 4,326 questions \\
HotpotQA & Offline localization and evidence analysis & 100 sampled trajectories \\
\bottomrule
\end{tabular}
\end{table}

\paragraph{Unified answer evaluation.}
We use the GPT-5.4 API as a common judge across all benchmarks, models, and
methods. Given the question, reference answer, and model's final response,
the judge assesses semantic correctness, accepting equivalent wording.
Success is the fraction of evaluated questions whose final answers are
judged correct. Trajectories that terminate with a tool error, timeout,
unparseable answer, or exhausted interaction budget are counted as failures.
They remain in the success-rate denominator and contribute their full
elapsed time to the reported mean latency.

\paragraph{Head-localization samples.}
For each model, we randomly sample 100 HotpotQA trajectories using seed 42
and select one input containing correct-answer tokens from each trajectory.
Localization uses the final query token's attention at the end of prefill,
as detailed in Appendix~\ref{app:implementation_details}. The resulting
head set remains fixed for all subsequent evaluations on the other datasets.

\paragraph{Evidence-ranking evaluation.}
For the BrowseComp-Plus scoring analysis in
Section~\ref{sec:signal_validation}, candidate sets are drawn from the same
search pipeline used during agent execution. For each evaluation query, we
take the first 5 or 20 chunks from a shared search-result list in their
original retrieval order. This keeps the scoring evaluation close to the
retrieval conditions encountered during reasoning. Positive chunks are
identified using dataset-provided evidence annotations where available and
semantic judgments of whether a chunk supports the answer otherwise.
All models use the same candidates and relevance labels.

We compute AUROC and Recall@$k$ within each query's candidate set and report
macro-averages across queries. Recall@$k$ is the fraction of positive chunks
retained among the top $k$ scored candidates, with $k\in\{1,3\}$ for the
5-chunk setting and $k\in\{1,10\}$ for the 20-chunk setting. Each metric is
averaged over the queries for which it is defined: AUROC requires both
positive and negative candidates, while Recall@$k$ requires at least one
positive candidate. Equal scores in top-$k$ selection retain the original
retrieval order.

\subsection{Models and Agent Settings}
\label{app:model_agent_settings}

DeepDiver-7B is the primary model for the four-benchmark comparison.
OpenResearcher-30B and OpenPangu-2.0-Flash-92B additionally support the
BrowseComp-Plus cross-model comparison. LiteResearcher-4B and WebSeer-14B
appear in the signal analysis. Table~\ref{tab:model_revisions} lists the
official checkpoints. We follow each model's released framework for model
loading, system prompts, chat templates, and tool definitions. The agent
budgets used in our experiments are listed in
Table~\ref{tab:agent_configuration}.

\begin{table}[H]
\centering
\caption{Official checkpoint identifiers, linked to their model cards.}
\label{tab:model_revisions}
\normalsize
\setlength{\tabcolsep}{4pt}
\renewcommand{\arraystretch}{1.12}
\begin{tabular}{@{}>{\raggedright\arraybackslash}p{0.32\textwidth}
                  >{\raggedright\arraybackslash}p{0.63\textwidth}@{}}
\toprule
Model & Official checkpoint \\
\midrule
DeepDiver-7B & \href{https://huggingface.co/openpangu/openPangu-Embedded-7B-DeepDiver}{\nolinkurl{openpangu/openPangu-Embedded-7B-DeepDiver}} \\
OpenResearcher-30B & \href{https://huggingface.co/OpenResearcher/OpenResearcher-30B-A3B}{\nolinkurl{OpenResearcher/OpenResearcher-30B-A3B}} \\
OpenPangu-2.0-Flash-92B & \href{https://huggingface.co/openpangu/openPangu-2.0-Flash}{\nolinkurl{openpangu/openPangu-2.0-Flash}} \\
LiteResearcher-4B & \href{https://huggingface.co/simplex-ai-inc/LiteResearcher-4B}{\nolinkurl{simplex-ai-inc/LiteResearcher-4B}} \\
WebSeer-14B & \href{https://huggingface.co/99hgz/WebSeer-14b}{\nolinkurl{99hgz/WebSeer-14b}} \\
\bottomrule
\end{tabular}
\end{table}

\begin{table}[H]
\centering
\caption{Generation and interaction budgets for the end-to-end agents.
Within each model--benchmark setting, all compared methods use the same budgets.}
\label{tab:agent_configuration}
\normalsize
\renewcommand{\arraystretch}{1.12}
\setlength{\tabcolsep}{4pt}
\begin{tabular}{@{}>{\raggedright\arraybackslash}p{0.35\textwidth}
                  >{\raggedright\arraybackslash}p{0.32\textwidth}
                  >{\raggedright\arraybackslash}p{0.27\textwidth}@{}}
\toprule
Model & Max. output tokens per call & Max. interaction rounds \\
\midrule
DeepDiver-7B & 256 & 50 \\
OpenResearcher-30B & 8,192 & 50 \\
OpenPangu-2.0-Flash-92B & 4,096 & 50 \\
\bottomrule
\end{tabular}
\end{table}

\paragraph{Decoding and termination.}
Across the end-to-end experiments, we use temperature $0.5$, top-$p=0.95$,
and top-$k=20$. Each trajectory is limited to 50 interaction rounds and
terminates when the agent produces a final answer or reaches this budget.
For OpenPangu, two of the 50 model requests are reserved for producing the
final answer. The per-call output limits are 256 tokens for DeepDiver,
8,192 for OpenResearcher, and 4,096 for OpenPangu.

\paragraph{Context limits and error handling.}
When an input exceeds the model framework's context limit, we discard the
oldest context entries and retain the most recent content until the input
fits. Malformed tool calls are allowed one fallback attempt. Timeout
protection and bounded retries handle transient request and generation
failures. If recovery fails, the trajectory terminates and is counted as
unsuccessful, with its elapsed time included as specified in
Appendix~\ref{app:benchmark_protocol}. These safeguards apply consistently
to all methods within each experimental setting.

\paragraph{QRAE evaluator.}
The evaluator uses the same frozen reasoning model as the agent and performs
prefill only, generating no output tokens. Its input serialization,
attention-token ranges, and cross-step prompt updates are specified in
Appendix~\ref{app:implementation_details}.

\subsection{Retrieval and Compression Settings}
\label{app:retrieval_compression_settings}

BrowseComp-Plus uses its fixed local corpus, while XBench-DeepSearch,
FRAMES, and SimpleQA use online search through Serper.
A chunk is one search-returned text snippet, constructed using the limits in
Table~\ref{tab:retrieval_configuration}. \MethodName adds no further splitting
or overlap. It compresses only \texttt{search} results; \texttt{open} and
\texttt{find} observations are passed to the agent without compression.
The scoring batch limit $K$ equals the number of results requested per query,
and the default retention rate is $\rho=0.5$.

\begin{table}[H]
\centering
\caption{Retrieval settings. All result counts are per query. Token and character
limits apply to snippet text before compression.}
\label{tab:retrieval_configuration}
\normalsize
\setlength{\tabcolsep}{4pt}
\renewcommand{\arraystretch}{1.15}
\begin{tabular}{@{}>{\raggedright\arraybackslash}p{0.20\textwidth}
                  >{\raggedright\arraybackslash}p{0.27\textwidth}
                  >{\raggedright\arraybackslash}p{0.10\textwidth}
                  >{\raggedright\arraybackslash}p{0.35\textwidth}@{}}
\toprule
Setting & Retrieval source & Top-$k$ & Search snippet \\
\midrule
BrowseComp-Plus: DeepDiver & BM25, official index & 5 & First 512 tokens per document; Qwen3-0.6B tokenizer \\
BrowseComp-Plus: OpenResearcher & Official dense-retrieval service & 10 & Query-highlighted summary, at most 180 characters \\
BrowseComp-Plus: OpenPangu & BM25, official index & 10 & At most 256 tokens per document \\
XBench-DeepSearch & Serper web search & 10 & API snippet, truncated to 512 tokens if longer \\
FRAMES & Serper web search & 10 & API snippet, truncated to 512 tokens if longer \\
SimpleQA & Serper web search & 10 & API snippet, truncated to 512 tokens if longer \\
\bottomrule
\end{tabular}
\end{table}

\paragraph{Retrieval and snippet construction.}
DeepDiver and OpenPangu use BM25 over the official BrowseComp-Plus index.
DeepDiver follows the standard scaffold with five hits per query and the
first 512 tokens of each document, using the
\nolinkurl{Qwen/Qwen3-0.6B} tokenizer.\footnote{Official
\href{https://github.com/texttron/BrowseComp-Plus/blob/046949032b0328319cc9a02663a759ec601d9402/searcher/mcp_server.py}{server defaults} and
\href{https://github.com/texttron/BrowseComp-Plus/blob/046949032b0328319cc9a02663a759ec601d9402/searcher/tools.py}{snippet construction}.}
OpenResearcher uses
\nolinkurl{Qwen/Qwen3-Embedding-8B} with the corresponding prebuilt
BrowseComp-Plus dense index and a retrieval limit of ten results per query.
Its service highlights query-matching passages and truncates the summary to
180 characters, falling back to the first 180 characters when highlighting
fails.\footnote{Official
\href{https://github.com/TIGER-AI-Lab/OpenResearcher/blob/785fd6ba5fcbc068daa4a2f07bbe0964f2983c86/scripts/start_search_service.sh}{retriever launcher} and
\href{https://github.com/TIGER-AI-Lab/OpenResearcher/blob/785fd6ba5fcbc068daa4a2f07bbe0964f2983c86/scripts/deploy_search_service.py}{summary implementation}.}
For online search, we use the snippets returned directly by Serper and
truncate only those exceeding 512 tokens, keeping their leading text.
Online truncation and OpenPangu's local 256-token limit use the corresponding
agent tokenizer. Search results are deduplicated by URL; retained snippets
keep their associated source identifiers, titles, and URLs.

\paragraph{OpenPangu browsing configuration.}
Each \texttt{search} call accepts at most four queries and returns up to ten
documents per query. On BrowseComp-Plus, the snippet window is 256 tokens
per document; online search follows the Serper settings above.
The default \texttt{open} view starts at line 0 and returns 20 lines;
both the starting line and the requested line count can be changed.
\texttt{find} returns at most 20 distinct matching display lines, retaining
their original line numbers. Document text is wrapped at approximately
120 characters per display line, and both tools operate on those display
lines. These browsing observations remain uncompressed.
For a multi-query search call, \MethodName processes the result groups
separately in query order, updating the shared candidate cache after each
group. Each query therefore has its own scoring update with $K=10$ and
$M=5$; the results of several queries are not merged into a larger scoring batch.

\paragraph{QRAE scheduling and retention.}
When the candidate cache reaches $K$ chunks, at most one batch of $K$ chunks
is scored per query-level search update, prioritizing chunks with the fewest
prior scores. After their
historical arithmetic means are updated, the top
$M=\lceil\rho K\rceil$ scored chunks in the entire candidate cache enter the
persistent whitelist. Thus $K=5$ admits three chunks and $K=10$ admits five
at $\rho=0.5$; admission uses ranking rather than a score threshold.
Unscored chunks are ineligible for admission. Admitted chunks leave the
candidate cache and remain in the agent context. Ties in scoring priority
or mean score are resolved by first-seen retrieval order, which is also
preserved when retained snippets are presented to the agent.
The model-specific frozen head counts are 2, 3, and 5 for DeepDiver,
OpenResearcher, and OpenPangu, respectively (Table~\ref{tab:selected_heads}).

\paragraph{Compression baselines and budgets.}
Within each model--benchmark setting, all methods share the reasoning model,
retrieval configuration, and agent budgets in
Appendix~\ref{app:model_agent_settings}. We adapt the baseline interfaces to
the same search-result stream and protect system instructions, questions,
agent-generated reasoning, and non-search tool observations.
Table~\ref{tab:baseline_configuration} specifies the compression settings,
following the mechanisms in the respective
papers~\citep{pan2024llmlingua2,kang2026acon,sharifymoghaddam2026rerank,
zhang2025qrhead,kariyappa2026sidequest}.
We align retention with \MethodName as closely as each method permits:
passage selectors keep $\lceil K/2\rceil$ snippets per query, and text
compressors target half of the eligible snippet tokens. Source metadata is
preserved outside the text-compression budget. Since passage selection,
rewriting, and adaptive cache eviction operate at different granularities,
these targets do not imply identical realized token counts.

\begin{table}[H]
\centering
\caption{Compression configurations under matched agent and retrieval budgets.
Retention targets apply to search-derived content.}
\label{tab:baseline_configuration}
\normalsize
\renewcommand{\arraystretch}{1.13}
\begin{tabular}{@{}>{\raggedright\arraybackslash}p{0.21\textwidth}
                  >{\raggedright\arraybackslash}p{0.74\textwidth}@{}}
\toprule
Method & Configuration \\
\midrule
DeepSearch & Uncompressed retrieved context under the matched agent and retrieval budgets. \\
\MethodName & $\rho=0.5$ by default; fixed heads in Table~\ref{tab:selected_heads};
Search-aligned $K$ and $M=\lceil K/2\rceil$; online paired prefills;
no evaluator generation. \\
LLMLingua-2 & \nolinkurl{microsoft/llmlingua-2-xlm-roberta-large-meetingbank};
token retention rate 0.5 on each query's returned snippet text, before it
enters the agent context; source metadata and snippet boundaries are protected. \\
ACON & Observation-compression variant with GPT-4.1-mini and the published
observation-compression guideline, conditioned on the task and interaction
history. Compress a query's search observation when its snippet text exceeds
400 tokens, targeting 50\% of its original text length; retain only the refined
observation and leave the existing history unchanged. \\
Rerank & \nolinkurl{Qwen/Qwen3-Reranker-0.6B}, using its query--document
relevance prompt and yes/no score. Rerank the same $K$ retrieved snippets and
keep the top $\lceil K/2\rceil$; use the snippet limits above without retrieving
additional candidates. \\
QRHead & QRRetriever scoring on the corresponding frozen reasoning model:
16 heads for DeepDiver and 32 for OpenResearcher and OpenPangu. Average
query-token attention mass over selected heads, subtract the score for the
null query \texttt{N/A}, and retain the top $\lceil K/2\rceil$ snippets per query. \\
SideQuest & Adapt the auxiliary memory-management procedure to the corresponding
frozen reasoning model, invoked every four search updates. Select obsolete
search-result cursors for removal from context and KV state, with a soft
50\% retention target for eligible search-derived tokens; keep management
reasoning outside the main context and preserve non-search content. \\
\bottomrule
\end{tabular}
\end{table}

For QRHead, head selection uses the same 100 HotpotQA calibration inputs
specified in Appendix~\ref{app:benchmark_protocol}, with its own
query-to-relevant-passage attention criterion rather than \MethodName's
answer-token criterion. The selected heads are then fixed across benchmarks.
ACON's compression instruction preserves evidence needed for the remaining
task, including entities, dates, relations, and source attribution; it
requests the half-length target without adding unsupported facts.
SideQuest retains its model-driven deletion decisions: the 50\% target guides
memory-management requests, while the resulting cache size can vary with
the evidence still needed by the agent. These baseline adaptations preserve
the reasoning checkpoints and require no additional agent fine-tuning.

\subsection{Serving Environment and Cost Accounting}
\label{app:serving_cost_protocol}

Table~\ref{tab:serving_configuration} summarizes the hardware for the main
reasoning models. All models use BF16; the additional head-analysis models
use the same four-RTX-4090 configuration as the non-OpenPangu backbones.
Model parallelism configurations vary across models to accommodate
their architectures and resource requirements. All methods are run serially.
Within each model--benchmark
setting, \MethodName and the baselines use matched hardware allocations,
precision, backend settings, model parallelism, and agent budgets
for a fair comparison.

\begin{table}[H]
\centering
\caption{Hardware and precision for the end-to-end comparisons. Execution
settings are matched across methods within each model--benchmark setting.}
\label{tab:serving_configuration}
\normalsize
\setlength{\tabcolsep}{4pt}
\renewcommand{\arraystretch}{1.15}
\begin{tabular}{@{}>{\raggedright\arraybackslash}p{0.32\textwidth}
                  >{\raggedright\arraybackslash}p{0.44\textwidth}
                  >{\raggedright\arraybackslash}p{0.18\textwidth}@{}}
\toprule
Model & Hardware & Precision \\
\midrule
DeepDiver-7B & $4\times$ NVIDIA RTX 4090 & BF16 \\
OpenResearcher-30B & $4\times$ NVIDIA RTX 4090 & BF16 \\
OpenPangu-2.0-Flash-92B & $4\times$ NVIDIA H200 & BF16 \\
\bottomrule
\end{tabular}
\end{table}

\paragraph{Prefix reuse and attention access.}
The paired evaluator calls reuse their common prefix through vLLM prefix
caching. In the DeepDiver client, hooks at the selected layers capture the
final input token's query vector and the candidate tokens' key vectors after
positional encoding. Scaled query--key products reconstruct the selected-head
attention rows, normalized over candidate tokens in FP32. The query-conditioned
pass populates the prefix cache and stores the candidate key vectors; the
query-free pass reuses the prefix and captures its own final-token query
vector. This permits attention readout without recomputing the cached document
keys. Eager execution ensures that the hooks run while preserving the model's
original attention computation. The client invokes the backend with a one-token
output limit and discards the sampled output; scoring uses only prefill
activations.

\paragraph{Token accounting.}
Token counts are accumulated from backend-reported usage over the full
trajectory. Unless a table states otherwise, totals include input and output
tokens from agent calls, both \MethodName scoring calls, and any auxiliary
compression, reranking, or memory-management calls used by the baselines.
Reused prefixes remain included in reported input-token counts; cache reuse
reduces computation rather than removing these tokens from the accounting.
The OpenPangu columns in Table~\ref{tab:cross_model_end_to_end} report input
tokens only, with the same accounting convention applied to all methods.

\paragraph{Timing and aggregation.}
End-to-end time includes agent inference, retrieval and webpage processing,
and all online compression overhead. In particular, \MethodName's attention
scoring and context-management time are included, as are the corresponding
auxiliary costs of every baseline. We apply the same timing and aggregation
procedure to all methods within each setting and report means over all
evaluation trajectories. Failed trajectories remain included, as specified
in Appendix~\ref{app:benchmark_protocol}.

\subsection{Repeated-Run Protocol}
\label{app:run_protocol}

We use five seeds, $\{41,42,43,44,45\}$, for each benchmark, with the same
seed paired between DeepSearch and \MethodName in each run. Every run covers
all evaluation trajectories. Model generation and sampling settings remain
unchanged across runs, as specified in Appendix~\ref{app:model_agent_settings}.
The main comparison in Table~\ref{tab:end_to_end} uses seed 42.
Appendix~\ref{app:run_stability} reports the repeated-run results and their
stability analysis.

\subsection{Deferred Admission and Evidence Recovery}
\label{app:deferred_recovery}

We examine whether the candidate cache recovers evidence that is not admitted
immediately. A delayed admission occurs when a chunk is excluded from the
whitelist after its first retrieval but enters the whitelist at a later
step. We analyze the original compression runs on BrowseComp-Plus
(830 trajectories) and XBench-DeepSearch (100 trajectories), recording both
all delayed admissions and the annotated evidence among them.
Table~\ref{tab:deferred_recovery} summarizes the results. Chunk counts are
averaged over all trajectories; admission delay is measured from first
retrieval to later admission using the recorded trajectory-step indices.

% Statistics supplied by the author from the original compression runs:
% BrowseComp 20260706_001741 (830 trajectories), XBench 20260713_010030 (100).
% Delayed annotated evidence per trajectory: 1.31 and 2.69, respectively.
\begin{table}[H]
\centering
\caption{Delayed admission and recovery of annotated evidence in the original
compression runs. Evidence counts are a subset of all delayed admissions.}
\label{tab:deferred_recovery}
\normalsize
\setlength{\tabcolsep}{4pt}
\renewcommand{\arraystretch}{1.15}
\begin{tabular*}{\textwidth}{@{\extracolsep{\fill}}>{\raggedright\arraybackslash}p{0.52\textwidth}cc@{}}
\toprule
Metric & BrowseComp-Plus & XBench-DeepSearch \\
\midrule
Delayed admissions per trajectory & 4.36 & 11.17 \\
Annotated evidence among delayed admissions per trajectory & 1.31 & 2.69 \\
Mean admission delay (steps) & 2.00 & 2.10 \\
Trajectories with delayed admission & 745/830 (89.8\%) & 89/100 (89.0\%) \\
\bottomrule
\end{tabular*}
\end{table}

Delayed admission occurs in approximately nine out of ten trajectories on
both benchmarks, with a mean admission delay of about two steps. It recovers an average of
1.31 annotated evidence chunks per BrowseComp-Plus trajectory and 2.69 per
XBench-DeepSearch trajectory. These observations show that the candidate cache
is actively used to bring initially excluded evidence into the reasoning
context during ordinary execution.

In most cases of initially deferred correct snippets, other correct snippets
from the same search receive higher scores and take the available admission
slots. A useful snippet can therefore rank too low for immediate admission
when several relevant results compete under a limited retention budget.
The candidate cache is designed to handle this situation: it preserves the
remaining snippets outside the active reasoning context and makes them
available for subsequent reassessment and admission as candidate batches
and historical mean scores are updated.

Table~\ref{tab:deferred_recovery_cases} illustrates this process with two
XBench-DeepSearch trajectories. In question 66, a roast-duck snippet is
initially ranked sixth by both selected heads at step 3. It remains in the
candidate cache until step 5, when its head-wise ranks become second and
fifth and it enters the whitelist; the final answer is Peking duck and is
correct. In question 75, a snippet about \emph{The Voice of China} is deferred
at step 5 with ranks 14 and 12, then admitted at step 8 with ranks 2 and 1.
The final answer, Li Wei, is also correct. These cases trace how initially
lower-ranked evidence remains recoverable across steps and later becomes
available to the agent.

% Case details and answer correctness supplied by the author.
\begin{table}[H]
\centering
\caption{Two XBench-DeepSearch examples of evidence deferred and later
admitted. Rank pairs give the respective ranks assigned by the two selected
heads; smaller ranks are better.}
\label{tab:deferred_recovery_cases}
\normalsize
\setlength{\tabcolsep}{3pt}
\renewcommand{\arraystretch}{1.15}
\begin{tabular}{@{}>{\raggedright\arraybackslash}p{0.23\textwidth}
                  >{\raggedright\arraybackslash}p{0.23\textwidth}
                  >{\raggedright\arraybackslash}p{0.28\textwidth}
                  >{\raggedright\arraybackslash}p{0.20\textwidth}@{}}
\toprule
Example & Initially deferred & Reassessed and admitted & Final answer \\
\midrule
Question 66: roast duck & Step 3; ranks $(6,6)$ & Step 5; ranks $(2,5)$ & Peking duck (correct) \\
Question 75: \emph{The Voice of China} & Step 5; ranks $(14,12)$ & Step 8; ranks $(2,1)$ & Li Wei (correct) \\
\bottomrule
\end{tabular}
\end{table}

\section{Additional Experimental Results}
\label{app:additional_results}

\subsection{Cross-Model Evidence-Assessment Heads}
\label{app:cross_model_heads}

\begin{figure}[H]
  \centering
  \captionsetup[subfigure]{skip=2pt}
  \begin{subfigure}[b]{0.49\textwidth}
    \centering
    \includegraphics[width=\linewidth]{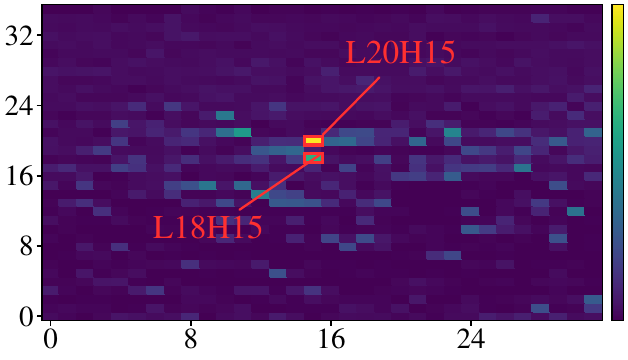}
    \caption{LiteResearcher-4B}
  \end{subfigure}\hfill
  \begin{subfigure}[b]{0.49\textwidth}
    \centering
    \includegraphics[width=\linewidth]{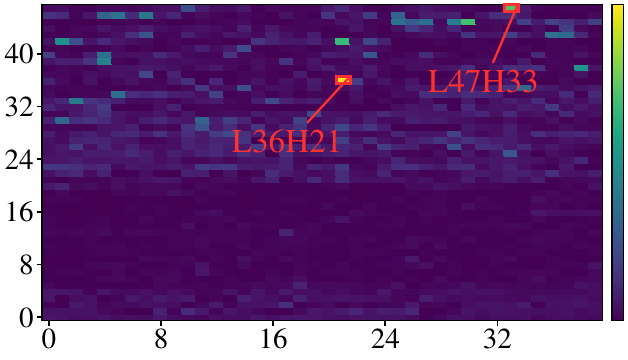}
    \caption{WebSeer-14B}
  \end{subfigure}
  \par\medskip
  \begin{subfigure}[b]{0.49\textwidth}
    \centering
    \includegraphics[width=\linewidth]{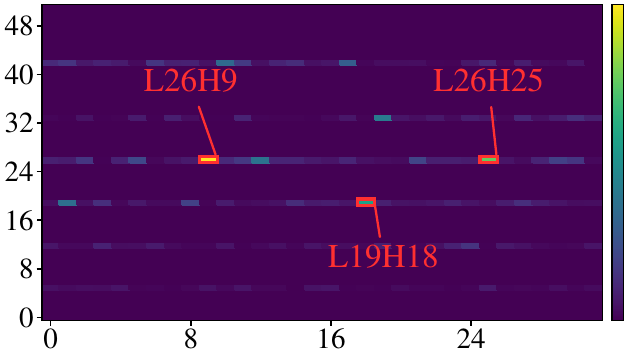}
    \caption{OpenResearcher-30B}
  \end{subfigure}\hfill
  \begin{subfigure}[b]{0.49\textwidth}
    \centering
    \includegraphics[width=\linewidth]{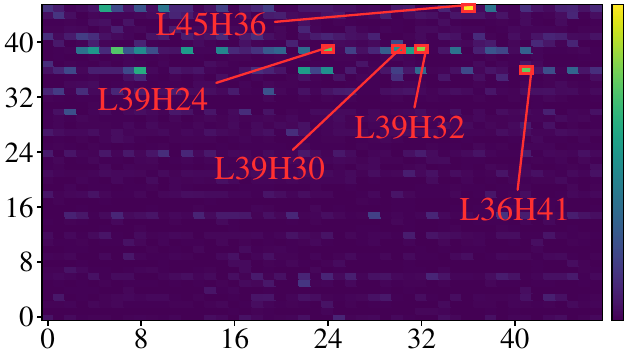}
    \caption{OpenPangu-2.0-Flash-92B}
  \end{subfigure}
  \caption{Cross-model \EAH localization. Red boxes mark the
  annotated heads. The horizontal and vertical axes denote head and layer
  indices, respectively, both starting from zero.}
  \label{fig:additional_model_heads}
  \label{fig:cross_architecture_heads}
\end{figure}

Figure~\ref{fig:additional_model_heads} shows head localization in
LiteResearcher-4B, WebSeer-14B, OpenResearcher-30B, and
OpenPangu-2.0-Flash-92B, complementing the DeepDiver results in
Figure~\ref{fig:head_localization}. All four models exhibit a small set of
high-response heads at model-specific locations. Examples include L20H15
and L18H15 in LiteResearcher, and L36H21 and L47H33 in WebSeer.

In OpenResearcher-30B's hybrid Mamba--Transformer architecture, which combines
linear-time sequence modeling with full attention~\citep{li2026openresearcher},
the localized \EAHs reside in the full-attention layers. In OpenPangu's
DeepSeek-style hybrid sparse attention~\citep{openpangu2026flash}, the signal
is concentrated in the main multi-head latent attention (MLA) heads rather
than the auxiliary indexer heads. These results support the presence of
\EAHs across the tested architectures, within the attention components
that aggregate contextual evidence.

\subsection{Robustness of QRAE Scoring}
\label{app:scoring_robustness}

We assess DeepDiver-7B's sensitivity to input-chunk order using the same
BrowseComp-Plus candidate sets and metrics as in
Table~\ref{tab:cross_model_ranking}. For both the 5- and 20-chunk settings,
we keep each query, its candidate chunks and evidence labels, and the
model's localized head set fixed, while randomly permuting the chunks
within each input. In each repetition, we rescore the full evaluation set
and compute AUC and each recall metric with the same aggregation as in
Table~\ref{tab:cross_model_ranking}. We summarize the variation across
repetitions with the sample STD of each metric.
% Specify the number of scoring-permutation repetitions for reproducibility.

\begin{table}[H]
\centering
\caption{Input-order sensitivity of \SignalName scoring with DeepDiver-7B
on BrowseComp-Plus. Entries show the reference score from
Table~\ref{tab:cross_model_ranking} $\pm$ STD across input permutations.
All scores use the percentage scale; STDs are in percentage points.}
\label{tab:scoring_order_robustness}
\normalsize
\setlength{\tabcolsep}{2pt}
\renewcommand{\arraystretch}{1.12}
\begin{tabular*}{\textwidth}{@{\extracolsep{\fill}}lcccccc@{}}
\toprule
\multirow{2}{*}{Model}
& \multicolumn{3}{c}{5 chunks}
& \multicolumn{3}{c}{20 chunks} \\
\cmidrule(lr){2-4}\cmidrule(lr){5-7}
& AUC & Recall@1 & Recall@3
& AUC & Recall@1 & Recall@10 \\
\midrule
% Reference scores are copied from Table 3, not estimated permutation means.
% Permutation STDs supplied by the author.
DeepDiver
& $83.2\pm0.5$ & $(54.6\pm0.4)\%$ & $(83.2\pm0.1)\%$
& $84.4\pm0.2$ & $(44.6\pm0.6)\%$ & $(88.7\pm0.1)\%$ \\
\bottomrule
\end{tabular*}
\end{table}

Across both candidate-set sizes, the STDs range from $0.1$ to $0.6$
percentage points. This low variation supports the stability of overall
evidence discrimination and top-$k$ evidence recall under input-order
perturbations, without requiring individual chunk scores to be numerically
identical.

\paragraph{Case study: \texttt{qid-483}.}
As a complementary case study on BrowseComp-Plus, we run \MethodName
20 times on \texttt{qid-483}, randomly permuting the retrieved chunks
after each search while holding the model, decoding seed, retrieval
configuration, and serving environment fixed. We enable deterministic
kernels with \texttt{BATCH\_INVARIANT=1}. All 20 runs produce identical
agent-generated trajectories and final answers. Both input- and output-token
counts have zero STD, while runtime has an STD of $0.1745$\,s. This case
illustrates consistent downstream behavior under the tested retrieval-order
permutations.

\subsection{Stability across Random Seeds}
\label{app:run_stability}

To assess the reliability of the end-to-end gains,
Table~\ref{tab:run_stability} summarizes five complete runs of DeepSearch and
\MethodName on each of the four benchmarks, following the paired-seed
protocol in Appendix~\ref{app:run_protocol}.
Online search uses the Serper API, whose returned results can also vary across
runs, adding retrieval variability to seed-induced variation.

For each run, success is the fraction of successful trajectories, while tokens
and latency are averaged over all trajectories, including unsuccessful ones.
End-to-end costs include evaluator overhead, as defined in
Section~\ref{sec:setup}. Entries report the mean and sample STD across the five
runs, measuring run-to-run variability rather than differences between
questions within one run.

% Values supplied by the author, who specifies five complete runs per benchmark.

\begin{table}[H]
\centering
\caption{Comparison of DeepSearch and \MethodName over five paired seeds (41--45) per benchmark,
with every run covering all evaluation trajectories. Entries show mean
$\pm$ sample STD; success is in percent, time in seconds, and tokens
in thousands.}
\label{tab:run_stability}
\normalsize
\setlength{\tabcolsep}{3pt}
\renewcommand{\arraystretch}{1.12}
\begin{tabular*}{\textwidth}{@{\extracolsep{\fill}}llccc@{}}
\toprule
Dataset & Method & Success (\%) $\uparrow$ & Time (s) $\downarrow$
& Tokens (k) $\downarrow$ \\
\midrule
\multirow{2}{*}{\shortstack[l]{BrowseComp-\\Plus}}
& DeepSearch & $13.7 \pm 0.5$ & $29.6 \pm 0.3$ & $177 \pm 19$ \\
& \textbf{\MethodName} & $\boldsymbol{17.4 \pm 0.3}$ & $\boldsymbol{20.6 \pm 0.1}$ & $\boldsymbol{55 \pm 4}$ \\
\midrule
\multirow{2}{*}{\shortstack[l]{XBench-\\DeepSearch}}
& DeepSearch & $38.4 \pm 3.1$ & $265.8 \pm 9.4$ & $311 \pm 24$ \\
& \textbf{\MethodName} & $\boldsymbol{44.6 \pm 1.8}$ & $\boldsymbol{152.3 \pm 7.7}$ & $\boldsymbol{120 \pm 15}$ \\
\midrule
\multirow{2}{*}{FRAMES}
& DeepSearch & $40.9 \pm 1.1$ & $160.2 \pm 5.6$ & $138 \pm 14$ \\
& \textbf{\MethodName} & $\boldsymbol{42.7 \pm 1.2}$ & $\boldsymbol{91.8 \pm 3.7}$ & $\boldsymbol{92 \pm 10}$ \\
\midrule
\multirow{2}{*}{SimpleQA}
& DeepSearch & $80.1 \pm 0.9$ & $143.1 \pm 7.7$ & $73 \pm 8$ \\
& \textbf{\MethodName} & $\boldsymbol{82.7 \pm 0.5}$ & $\boldsymbol{99.5 \pm 5.8}$ & $\boldsymbol{45 \pm 3}$ \\
\bottomrule
\end{tabular*}
\end{table}

Across the four benchmarks, \MethodName reduces tokens by 50.5\% and
end-to-end latency by 36.6\%, while increasing success by 3.58 percentage
points. These equally weighted averages use per-benchmark relative cost
reductions and absolute success differences, computed from run means.
Efficiency gains are large relative to the reported STDs; mean success improves
on all four benchmarks, with a smaller margin relative to variability on
FRAMES. These repeated-run results support a stable effectiveness--efficiency
improvement.

\subsection{Functional Role of Query-Relevant Heads}
\label{app:expanded_intervention}
\label{sec:causal_validation}

We test whether the localized heads play a functional role in preparing
evidence for subsequent reasoning. Using annotated evidence, we add a
controlled bias $\beta$ to these heads' attention logits and measure
downstream task success.
For chunk intervention, the same bias is added to every token in the annotated
gold chunk. Token intervention restricts the patch to its annotated evidence
span. ``Prefill'' removes the patch before autoregressive decoding, whereas
``all'' retains it throughout generation. All intervention strengths are
evaluated over the same examples.

\begin{table}[H]
\centering
\caption{Success (\%) under attention intervention. ``Prefill'' applies the
patch only during prefill; ``all'' retains it through decoding.}
\label{tab:intervention}
\normalsize
\begin{tabular}{llrrrrrr}
\toprule
Target & Scope & $\beta=0$ & 2 & 3 & 4 & 5 & 6 \\
\midrule
Chunk & Prefill & 11 & 11 & 12 & 15 & 16 & 19 \\
Chunk & All     & 11 & 11 & 12 & 14 & 16 & 19 \\
Token & Prefill & 11 & 11 & 12 & 12 & 14 & 15 \\
Token & All     & 11 & 11 & 12 & 13 & 14 & 16 \\
\bottomrule
\end{tabular}
\end{table}

Table~\ref{tab:intervention} shows that chunk-level intervention raises
success from 11\% at $\beta=0$ to 19\% at $\beta=6$. At higher bias strengths,
chunk-level intervention is more effective than token-level intervention,
suggesting a benefit from highlighting the full evidence context.
Prefill-only intervention achieves gains comparable to retaining the patch
through decoding. Together, these results support the view that the localized
heads help prioritize relevant evidence during prefill for subsequent reasoning.

\subsection{Time and Token Cost Breakdown}
\label{app:cost_breakdown}

We analyze two representative benchmarks: BrowseComp-Plus with local
retrieval and XBench-DeepSearch with online web search.
Table~\ref{tab:cost_breakdown} separates per-trajectory time into reasoning,
retrieval, and compression, and token usage into agent inputs, agent outputs,
and compression inputs. This breakdown captures both the additional cost of
assessment and the savings in subsequent reasoning and tool use.

% Cost components restored from 9e51f3d:paper-token.tex (2026-07-22).
% Search-call and cache observations below are supplied by the author.
\begin{table}[H]
\centering
\caption{Per-trajectory time and token breakdown for DeepSearch and
\MethodName. Time is measured in seconds and token counts in thousands (k).
Input and output refer to the reasoning agent; compression costs are listed
separately. Dashes indicate no compression component.}
\label{tab:cost_breakdown}
\normalsize
\setlength{\tabcolsep}{3pt}
\renewcommand{\arraystretch}{1.12}
\begin{tabular*}{\textwidth}{@{\extracolsep{\fill}}llrrrrrr@{}}
\toprule
\multirow{2}{*}{Dataset} & \multirow{2}{*}{Method}
& \multicolumn{3}{c}{Time (s)} & \multicolumn{3}{c}{Tokens (k)} \\
\cmidrule(lr){3-5}\cmidrule(lr){6-8}
& & Reasoning & Tool & Compression & Input & Output & Compression \\
\midrule
\multirow{2}{*}{\shortstack[l]{BrowseComp-\\Plus}}
& DeepSearch & 28.2 & 1.3 & -- & 175 & 2.4 & -- \\
& \MethodName & 17.3 & 0.9 & 2.3 & 48 & 1.3 & 4.9 \\
\midrule
\multirow{2}{*}{\shortstack[l]{XBench-\\DeepSearch}}
& DeepSearch & 144.3 & 131.7 & -- & 314 & 4.6 & -- \\
& \MethodName & 41.1 & 98.6 & 10.7 & 77 & 2.8 & 37 \\
\bottomrule
\end{tabular*}
\end{table}

\SignalName scoring reads prefill activations without generating a reasoning
sequence. Each call processes a bounded batch of search snippets rather than
the full accumulated trajectory, keeping its input substantially shorter
than the reasoning context in long trajectories. The resulting compression
time is modest: 2.3\,s on BrowseComp-Plus and 10.7\,s on XBench-DeepSearch,
compared with reasoning-time savings of 10.9\,s and 103.2\,s, respectively.
Reasoning time falls by approximately 39\% and 72\%; including compression
and retrieval, end-to-end time still falls by approximately 31\% and 46\%.

The gains also reflect changes in search behavior. We observe approximately
30\% fewer search calls with \MethodName. Shorter contexts with fewer
irrelevant results can make the retained evidence easier to use, particularly
on difficult, long trajectories, and help the agent formulate more targeted
follow-up search queries. This provides a plausible explanation for reaching
an answer with fewer searches. The reduction matters especially for online
retrieval: on XBench-DeepSearch, retrieval time falls from 131.7\,s to
98.6\,s, saving 33.1\,s per trajectory.

Token savings arise from both admitting less search text into the reasoning
context and taking fewer search and reasoning steps. Agent input tokens fall
by approximately 73\% on BrowseComp-Plus and 75\% on XBench-DeepSearch, while output tokens fall by
approximately 46\% on BrowseComp-Plus and 39\% on XBench-DeepSearch. Including
the additional evaluator inputs, total token usage remains approximately
54k and 117k, compared with 177k and 319k for DeepSearch.

Evaluation introduces additional uncached input tokens when candidate batches
are first processed, requiring fresh prefill computation. These inputs are
therefore more costly to process than an already cached prefix. Nevertheless,
in the evaluated runs, the combined uncached input volume of evaluation and
reasoning remains below the uncached reasoning input volume of DeepSearch.
Cached input volume and output tokens also decrease substantially. Thus,
the extra evaluation inputs are offset even when cached and uncached inputs
are distinguished, rather than treated as equally costly.

\MethodName retains the reasoning-and-tool loop and the backend's KV-cache
semantics, changing which search results enter the active context. Prefix
reuse remains enabled, including reuse between the paired evaluator calls,
and the observed cache-hit rate is nearly unchanged. The savings therefore
come from processing less context and taking fewer calls while retaining
prefix reuse, rather than from a higher cache-hit rate.

A further deployment option is to expose a smaller evaluator through an API
while using a larger model for reasoning. The evaluator could score candidate
batches asynchronously alongside independent reasoning or tool work, reducing
assessment cost and potentially hiding part of its latency. This is a
prospective deployment extension: its net benefit would depend on evidence
selection quality, available overlap, and communication overhead, and is not
included in the reported gains.

\subsection{Effect of Head Selection}
\label{app:head_selection_ablation}

We isolate the effect of head selection while keeping query-free subtraction
and normalized-entropy scoring fixed. All three conditions use two heads on
DeepDiver-7B and the same BrowseComp-Plus candidate sets, evidence labels, and
retained-chunk budgets as Table~\ref{tab:scoring_ablations}. We compare the
heads localized by \MethodName with randomly selected heads and heads selected
by QRHead, applying the same \SignalName scoring rule to each set.

% Values supplied by the author; no standard deviations were provided.
\begin{table}[H]
\centering
\caption{Head-selection comparison on BrowseComp-Plus with DeepDiver-7B.
All head sets use the same \SignalName scoring rule. AUC denotes AUROC;
all metrics are percentages and higher is better.}
\label{tab:head_selection_ablation}
\normalsize
\setlength{\tabcolsep}{2pt}
\renewcommand{\arraystretch}{1.12}
\begin{tabular*}{\textwidth}{@{\extracolsep{\fill}}lcccccc@{}}
\toprule
\multirow{2}{*}{Head set}
& \multicolumn{3}{c}{5 chunks}
& \multicolumn{3}{c}{20 chunks} \\
\cmidrule(lr){2-4}\cmidrule(lr){5-7}
& AUC & Recall@1 & Recall@3
& AUC & Recall@1 & Recall@10 \\
\midrule
\textbf{\MethodName\ (full)} & \textbf{83.2} & \textbf{54.6\%} & \textbf{83.2\%} & \textbf{84.4} & \textbf{44.6\%} & \textbf{88.7\%} \\
Random heads & 45.7 & 9.9\% & 38.8\% & 53.0 & 5.3\% & 40.2\% \\
Heads selected by QRHead & 68.3 & 24.5\% & 70.0\% & 70.4 & 13.9\% & 63.3\% \\
\bottomrule
\end{tabular*}
\end{table}

Random heads provide little useful evidence discrimination: their AUCs of
45.7 and 53.0 are close to chance, and Recall@1 is only 9.9\% and 5.3\% in the
5- and 20-chunk settings, respectively. Heads selected by QRHead show a
stronger association with evidence, reaching AUCs of 68.3 and 70.4, but remain
consistently weaker than the heads localized by \MethodName. The full method
improves AUC over this control by 14.9 and 14.0 percentage points and achieves
higher recall on every reported metric. With the head count and scoring rule
held fixed, these results support the value of \MethodName's offline
localization for extracting an effective evidence-ranking signal.

% Ranked QRHead localization results supplied by the author.
Applying the QRHead localization procedure to DeepDiver-7B yields the following
top-10 heads in descending rank: \texttt{L16H10}, \texttt{L14H31},
\texttt{L16H8}, \texttt{L12H3}, \texttt{L16H16}, \texttt{L12H27},
\texttt{L14H10}, \texttt{L15H6}, \texttt{L11H30}, and \texttt{L14H28}.
None overlaps with the heads localized by \MethodName, \texttt{L18H22} and
\texttt{L17H0}. Although both approaches identify heads responsive to
query-related content, they select different head sets in this model.
One possible interpretation is that QRHead selection emphasizes query
retrieval and the routing of matching context, whereas \MethodName localization
favors heads that gate the flow of evidence during prefill. This difference
in emphasis may explain why QRHead-selected heads retain some relevance
signal but provide weaker evidence discrimination under our scoring rule.

\section{Theoretical Analysis of QRAE}
\label{app:theory}

This appendix records conditional properties of the three components used by
\MethodName: offline head localization, online query-free control subtraction, and
entropy-based admission. It also formalizes the bounded evaluator input and the
prompt-token savings from deferred propagation. These results explain the
construction of the method under explicit assumptions; they do not establish
that high-entropy response is semantically relevant or that compression
preserves task accuracy. Those claims are evaluated empirically in
Section~\ref{sec:experiments}.

\subsection{Finite-Sample Stability of Head Localization}

Let $\mathcal G$ be the set of candidate layer--head pairs and
$G=|\mathcal G|$. Write
$\mathcal D_{\mathrm{cal}}=\{(Q_n,d_n,E_n)\}_{n=1}^N$. For each
calibration input and head $g$, let
$Z_{n,g}=\sum_{u\in E_n}R_g(u;\mathcal X_n)\in[0,1]$ denote the
final-query-token attention mass on the labeled answer tokens, as used in
Equation~\ref{eq:head_localization}.
Define its population and empirical means as
\begin{equation*}
  \mu_g=\mathbb E[Z_{n,g}],
  \qquad
  \widehat\mu_g=\frac{1}{N}\sum_{n=1}^{N}Z_{n,g}.
\end{equation*}
We use $K_{\mathrm H}$ for the number of selected heads, distinguishing it from
the online candidate-batch limit $K$.

\begin{proposition}[Uniform calibration bound]
\label{prop:head_calibration}
If the calibration inputs are independent draws from a common calibration
distribution, then for any
$\delta\in(0,1)$, with probability at least $1-\delta$,
\begin{equation}
  \max_{g\in\mathcal G}|\widehat\mu_g-\mu_g|
  \leq
  \epsilon_N(\delta)
  :=\sqrt{\frac{\log(2G/\delta)}{2N}}.
  \label{eq:uniform_head_bound}
\end{equation}
If the gap between the $K_{\mathrm H}$-th and $(K_{\mathrm H}+1)$-th largest
population means is greater than $2\epsilon_N(\delta)$, empirical top-$K_{\mathrm H}$
selection recovers the population top-$K_{\mathrm H}$ set on the same event.
\end{proposition}

\begin{proof}
For a fixed head, Hoeffding's inequality gives
$\Pr(|\widehat\mu_g-\mu_g|>\epsilon)\leq2e^{-2N\epsilon^2}$
for variables in $[0,1]$.
A union bound over $G$ heads and the choice
$\epsilon=\epsilon_N(\delta)$ prove Equation~\ref{eq:uniform_head_bound}.
On this event, every empirical mean is within $\epsilon_N(\delta)$ of its
population value. A population boundary gap larger than twice this deviation
prevents any head below the boundary from overtaking a head above it.
\end{proof}

Proposition~\ref{prop:head_calibration} gives a sufficient condition for
repeatable offline localization. It does not assert that the required
population gap exists or that the same heads must appear in different models;
the cross-dataset and cross-model analyses test those empirical questions.

\subsection{Robustness of Matched-Control Subtraction}

Consider the normalized attention rows restricted to an evaluation batch.
Write
\begin{equation}
  \bar a^{(Q)}=b+\Delta+e_Q,
  \qquad
  \bar a^{(0)}=b+e_0,
  \label{eq:matched_decomposition}
\end{equation}
where $b$ is response shared by the query and control inputs, $\Delta$ is the
ideal query-induced change, and $e_Q,e_0$ collect mismatch and measurement
error. Let $x=[\bar a^{(Q)}-\bar a^{(0)}]_+$ be the \SignalName response and
$x^\star=[\Delta]_+$ its ideal counterpart.

\begin{lemma}[Matched-control robustness]
\label{lem:matched_control}
For any decomposition in Equation~\ref{eq:matched_decomposition},
\begin{equation*}
  \|x-x^\star\|_1\leq\|e_Q-e_0\|_1.
\end{equation*}
Thus the shared component $b$ cancels exactly, and control mismatch perturbs the
positive response by no more than its own $\ell_1$ discrepancy.
\end{lemma}

\begin{proof}
Subtraction cancels $b$ and yields
$x=[\Delta+e_Q-e_0]_+$. Coordinatewise positive clipping is 1-Lipschitz, so
$\|[u]_+-[v]_+\|_1\leq\|u-v\|_1$. Taking $u=\Delta+e_Q-e_0$ and
$v=\Delta$ proves the claim.
\end{proof}

The bound can be propagated through within-chunk normalization. Suppose
$\rho=\|x^\star\|_1>0$, $\eta=\|e_Q-e_0\|_1<\rho$, and let
$p=x/\|x\|_1$ and $p^\star=x^\star/\rho$. Then
\begin{equation}
  \|p-p^\star\|_1
  \leq \frac{2\eta}{\rho-\eta}.
  \label{eq:normalized_response_bound}
\end{equation}
Indeed, Lemma~\ref{lem:matched_control} implies
$\|x\|_1\geq\rho-\eta$; adding and subtracting
$x^\star/\|x\|_1$ proves Equation~\ref{eq:normalized_response_bound}.
Consequently, when the ideal positive mass is separated from zero, standard
entropy-continuity bounds also control the change in the normalized \SignalName score.
In particular, total variation satisfies
\begin{equation}
  \operatorname{TV}(p,p^\star)
  \leq \frac{\eta}{\rho-\eta},
  \label{eq:matched_control_tv}
\end{equation}
which supplies the perturbation radius used in Proposition
\ref{prop:robust_breadth} whenever the right-hand side satisfies that
proposition's range condition.

\subsection{Robust Entropy Separation}
\label{app:robust_entropy_separation}
\label{sec:response_breadth}

For a nonzero positive response on a chunk of length $n>1$, let $p$ denote
the within-chunk distribution obtained by normalizing
$\{\Delta_h(u)\}_{u\in d}$. Write
$E(p)=\Entropy(p)/\log n$, as in Equation~\ref{eq:qrae_entropy}.
Define $g_n(\delta)=h_2(\delta)+\delta\log(n-1)$, where $h_2$ is binary entropy.

\begin{proposition}[Effective response support]
\label{prop:effective_support}
For every distribution $p$ over $n$ tokens,
\begin{equation*}
  0\leq E(p)\leq1,
  \qquad
  \exp(\Entropy(p))=n^{E(p)}.
\end{equation*}
The lower bound is attained by a point mass and the upper bound by the uniform
distribution. The score is invariant to positive rescaling of the unnormalized
response $x$. If $p$ is uniform on exactly $m$ tokens, then
\begin{equation*}
  E(p)=\frac{\log m}{\log n}.
\end{equation*}
\end{proposition}

\begin{proof}
The entropy of a distribution on $n$ outcomes lies in $[0,\log n]$, with the
stated equality cases. Exponentiating $\Entropy(p)=E(p)\log n$ gives the effective
support identity. Positive rescaling cancels between the numerator and
denominator of $p_u=x_u/\sum_vx_v$. A uniform distribution on $m$ tokens has
entropy $\log m$.
\end{proof}

We now state the condition under which the constructed signal ranks an
evidence-bearing response above a narrow distractor.

\begin{proposition}[Robust query-induced breadth separation]
\label{prop:robust_breadth}
For one selected head, consider an evidence chunk and a distractor of equal
length $n>1$. Let $p_{\mathrm e}^{\star}$ and $p_{\mathrm d}^{\star}$ be their
ideal normalized positive query-induced responses, and let $p_{\mathrm e}$ and
$p_{\mathrm d}$ be the distributions recovered by the matched-control
construction. Suppose $p_{\mathrm e}^{\star}$ is uniform on $m$ tokens,
$p_{\mathrm d}^{\star}$ is supported on at most $r<m$ tokens, with
$\operatorname{TV}(p_c,p_c^{\star})\leq\delta$,
$c\in\{\mathrm e,\mathrm d\}$,
$0\leq\delta\leq1-\frac{1}{n}.$
Then
\begin{equation}
E(p_{\mathrm e})-E(p_{\mathrm d})
\geq
\big[ \log(m/r)-2g_n(\delta) \big] / \log n.
\label{eq:robust_breadth_margin}
\end{equation}
Consequently, the \SignalName score ranks the evidence response above the distractor whenever
$\log(m/r)>2g_n(\delta)$.
\end{proposition}

\begin{proof}
For distributions on $n$ outcomes whose total-variation distance is at most
$\delta\leq1-1/n$, the entropy-continuity inequality gives
\begin{equation*}
  |\Entropy(p)-\Entropy(p^\star)|
  \leq h_2(\delta)+\delta\log(n-1)
  =g_n(\delta).
\end{equation*}
Because $p_{\mathrm e}^{\star}$ is uniform on $m$ tokens,
$\Entropy(p_{\mathrm e}^{\star})=\log m$. Because
$p_{\mathrm d}^{\star}$ is supported on at most $r$ tokens,
$\Entropy(p_{\mathrm d}^{\star})\leq\log r$. Therefore,
\begin{align*}
  \Entropy(p_{\mathrm e})&\geq\log m-g_n(\delta),\\
  \Entropy(p_{\mathrm d})&\leq\log r+g_n(\delta).
\end{align*}
Subtracting these inequalities and dividing by $\log n$ proves Equation
\ref{eq:robust_breadth_margin}. If the condition holds for every selected head,
averaging the per-head scores preserves the average of their positive margins.
\end{proof}

Proposition~\ref{prop:robust_breadth} makes the construction's inductive bias
explicit. With an exact control ($\delta=0$), the score margin is at least
$\log(m/r)/\log n$: broader positive support wins independently of response
scale. With imperfect matching, the ranking is preserved when the breadth margin
dominates the estimation error.

Equation~\ref{eq:matched_control_tv} connects control mismatch to the
proposition: for each chunk, a mismatch of mass $\eta$ against ideal positive
mass $\rho$ produces a total-variation radius no larger than
$\eta/(\rho-\eta)$. Deviations from exact uniformity can likewise be included in
the total-variation radius relative to the corresponding ideal response.

These results formalize response breadth, not semantic support. Diffuse noise
can also have high entropy, so evidence overlap must be verified empirically.
They also expose a residual length effect: for fixed $m$, the normalized score
$\log m/\log n$ decreases with chunk length. Normalization places scores on a
common $[0,1]$ scale but does not make them fully length-invariant.

\subsection{Concentration of Repeated Scores}

\begin{proposition}[Running-mean concentration]
\label{prop:score_concentration}
Let $S_1,\ldots,S_R\in[0,1]$ be independent evaluations of one chunk with a
common mean $\mu$, and let $\bar S_R=R^{-1}\sum_{r=1}^{R}S_r$. For any
$\epsilon>0$,
\begin{equation*}
  \Pr(|\bar S_R-\mu|\geq\epsilon)\leq2e^{-2R\epsilon^2}.
\end{equation*}
\end{proposition}

\begin{proof}
Apply Hoeffding's inequality to independent variables bounded in $[0,1]$.
\end{proof}

This conditional result describes the running-mean estimates used to rank
cached candidates. The actual manager selects the top $\lceil\rho K\rceil$
scored candidates after each evaluation; the proposition does not guarantee
that this ranking identifies the most useful evidence. Batch composition and
trajectory state can change between evaluations, so the independent
common-mean model is an idealization rather than a guaranteed property.

\subsection{Evaluation Cost and Amortized Savings}
\label{app:evaluation_cost}
\label{sec:framework-efficiency}

\SignalName scoring uses the paired query and control prefills described in
Section~\ref{sec:qrae-signal}. Each evaluation adds input tokens, while keeping
deferred chunks outside the reasoning context avoids repeatedly forwarding
their tokens through later agent steps (Figure~\ref{fig:qrae_context_flow_framework}).
We account for these two terms below.

Let $L_I$ and $L_Q$ be the instruction and query lengths, and suppose every
candidate chunk contains at most $L$ tokens. Let
$\ell_t=\sum_{d\in\mathcal B_t}|d|$ be the candidate-token volume at trigger
$t$, and let $\mathcal T_{\mathrm{eval}}$ contain all evaluation triggers.

\begin{proposition}[Bounded evaluator input]
\label{prop:bounded_evaluator}
At one evaluation trigger, the total input-token volume of the matched query
and control prefills is at most
\begin{equation}
  2L_I+L_Q+2KL,
  \label{eq:evaluator_token_bound}
\end{equation}
independently of the number of chunks stored in the candidate cache.
\end{proposition}

\begin{proof}
The query prefill contains $L_I+L_Q+\sum_{d\in\mathcal B_t}|d|$ tokens and the
control prefill contains $L_I+\sum_{d\in\mathcal B_t}|d|$. Since
$|\mathcal B_t|\leq K$ and $|d|\leq L$, their sum is bounded by
Equation~\ref{eq:evaluator_token_bound}.
\end{proof}

Writing $E_t=2L_I+L_Q+2\ell_t$ for the input-token cost of one evaluation,
the cumulative evaluator input is
\begin{equation*}
  E_{\mathrm{eval}}=\sum_{t\in\mathcal T_{\mathrm{eval}}}E_t.
\end{equation*}
Proposition~\ref{prop:bounded_evaluator} shows that the per-trigger input bound
is independent of the total candidate-cache size $|\mathcal C_t|$.

\begin{proposition}[Deferred propagation savings]
\label{prop:propagation_savings}
Suppose a chunk $d$ is retrieved at step $t_0$ and an uncompressed agent would
include it in every subsequent prompt through step $T$. If \MethodName first admits it
at step $t_a\geq t_0$, deferral avoids
\begin{equation*}
  (t_a-t_0)|d|
\end{equation*}
prompt tokens relative to immediate forwarding. If the chunk is never admitted,
the saving is $(T-t_0)|d|$, under the convention that retrieval at step $t_0$
first affects the prompt for step $t_0+1$.
\end{proposition}

\begin{proof}
Immediate forwarding includes $d$ in the prompts for steps
$t_0+1,\ldots,t_a$, whereas deferred admission does not. There are
$t_a-t_0$ such prompts, each contributing $|d|$ tokens. If admission never
occurs, the same count extends through step $T$.
\end{proof}

Let $\mathcal D_{\mathrm{def}}$ be the set of deferred chunks, and let $h(d)$
count the subsequent reasoning contexts from which chunk $d$ is withheld.
Summing the per-chunk savings in Proposition~\ref{prop:propagation_savings}
gives
\begin{equation*}
  S_{\mathrm{prop}}=\sum_{d\in\mathcal D_{\mathrm{def}}}h(d)|d|.
\end{equation*}
Evaluation is amortized in input-token volume when
$S_{\mathrm{prop}}>E_{\mathrm{eval}}$. Longer deferral horizons favor
\MethodName, whereas frequent triggers or short trajectories reduce net savings.

This condition concerns input-token volume, not uniformly lower attention
FLOPs or latency, which also depend on sequence length, model size, backend
implementation, and prefix reuse. Token savings alone do not guarantee
preserved task accuracy. Appendix~\ref{app:serving_cost_protocol} describes
prefix reuse and the full token and latency accounting. The end-to-end
evaluation in Section~\ref{sec:end_to_end} measures the realized savings,
including evaluator overhead, and tests whether useful evidence is preserved
through task success.

\end{document}